\documentclass[]{bytedance_seed}

\usepackage[toc,page,header]{appendix}

\usepackage{minitoc}
\usepackage{enumitem}
\usepackage{pifont}
\usepackage{amssymb}
\usepackage{makecell}

\usepackage{xcolor}
\usepackage{booktabs}
\usepackage{hyperref}

\usepackage{booktabs}
\usepackage{hyperref}
\definecolor{sblue}{HTML}{b2cbe1} 
\definecolor{spurple}{HTML}{d9d2e9} 
\definecolor{sred}{HTML}{97261c} 
\definecolor{sgreen}{HTML}{266d4a}

\usepackage{graphicx}
\usepackage{subcaption}
\usepackage{array}

\usepackage{latexsym}
\usepackage{enumitem}
\usepackage{tabularx}
\usepackage{graphicx} 
\usepackage{booktabs}
\usepackage{multirow} 
\usepackage{colortbl} 
\usepackage{tabularx}
\usepackage{makecell}
\usepackage{booktabs}
\usepackage{listings}
\usepackage{longtable}

\usepackage{wrapfig}
\usepackage{caption}
\usepackage{xspace}
\usetikzlibrary{intersections}
\usepackage{paracol}
\usepackage{lipsum}
\usepackage{array}

\usepackage{tcolorbox}
\tcbuselibrary{breakable, skins}
\usepackage{tikz}

\newcommand{\redmark}{\textcolor{sred}{\ding{55}}}
\newcommand{\greencheck}{\textcolor{sgreen}{\ding{51}}}
\newtcolorbox[auto counter, number within=section]{Prompt}[2][]{%
  colback=white, % Background color (light sky blue)
  colframe=sblue!150, % Frame color (sky blue)
  width=\textwidth, % Box width equal to page width
  arc=3mm, 
  boxrule=0.8mm, % Border thickness
  title=\large #2, % Caption text with smaller font size
  breakable, % Supports page breaks
  fonttitle=\small, % Title font size
  fontupper=\footnotesize, % Content font size
  #1 % Additional options
}

\newtcolorbox[auto counter, number within=section]{QuestionCase}[2][]{%
  colback=white,
  colframe=spurple!150,
  width=\textwidth, % Box width equal to page width
  arc=3mm, % Sharp corners
  boxrule=0.8mm, % Border thickness
  title=\large #2, % Caption text with smaller font size
  breakable, % Supports page breaks
  fonttitle=\small, % Title font size
  fontupper=\footnotesize, % Content font size
  #1 % Additional options
}
\usepackage{graphicx}
\usepackage{float}
\usepackage{wrapfig}
\newcommand{\benchmark}{\textsc{FinSearchComp}\xspace}

\title{FinSearchComp: Towards a Realistic, Expert-Level Evaluation of Financial Search and Reasoning}

\affiliation[1]{ByteDance Seed}
\affiliation[2]{Columbia Business School}

\contribution{Full author list in Contributions}

\abstract{
Search has emerged as core infrastructure for LLM-based agents and is widely viewed as critical on the path toward more general intelligence. 
Finance is a particularly demanding proving ground: analysts routinely conduct complex, multi-step searches over time-sensitive, domain-specific data, making it ideal for assessing both search proficiency and knowledge-grounded reasoning. 
Yet no existing open financial datasets evaluate data searching capability of end-to-end agents, largely because constructing realistic, complicated tasks requires deep financial expertise and time-sensitive data is hard to evaluate. 
We present \benchmark, the first fully open-source agent benchmark for realistic, open-domain financial search and reasoning. 
\benchmark comprises three tasks---Time-Sensitive Data Fetching, Simple Historical Lookup, and Complex Historical Investigation---closely reproduce real-world financial analyst workflows. 
To ensure difficulty and reliability, we engage $70$ professional financial experts for annotation and implement a rigorous multi-stage quality-assurance pipeline. 
The benchmark includes $635$ questions spanning global and Greater China markets, and we evaluate $21$ models (products) on it. 
Grok 4 (web) tops the global subset, approaching expert-level accuracy. DouBao (web) leads on the Greater China subset. 
Experimental analyses show that equipping agents with web search and financial plugins substantially improves results on \benchmark, and the country origin of models and tools impact performance significantly.
By aligning with realistic analyst tasks and providing end-to-end evaluation, \benchmark offers a professional, high-difficulty testbed for complex financial search and reasoning.
}
\vspace{-0.2in}
\checkdata[Github]{\url{https://randomtutu.github.io/FinSearchComp/}}
\checkdata[Huggingface]{\url{ https://huggingface.co/ByteSeedXpert/FinSearchComp/}}
\begin{document}
\maketitle
\enlargethispage{\baselineskip}
\begin{figure}[h]
    \vspace{-0.3in}
    \centering
    \includegraphics[width=0.9\textwidth]{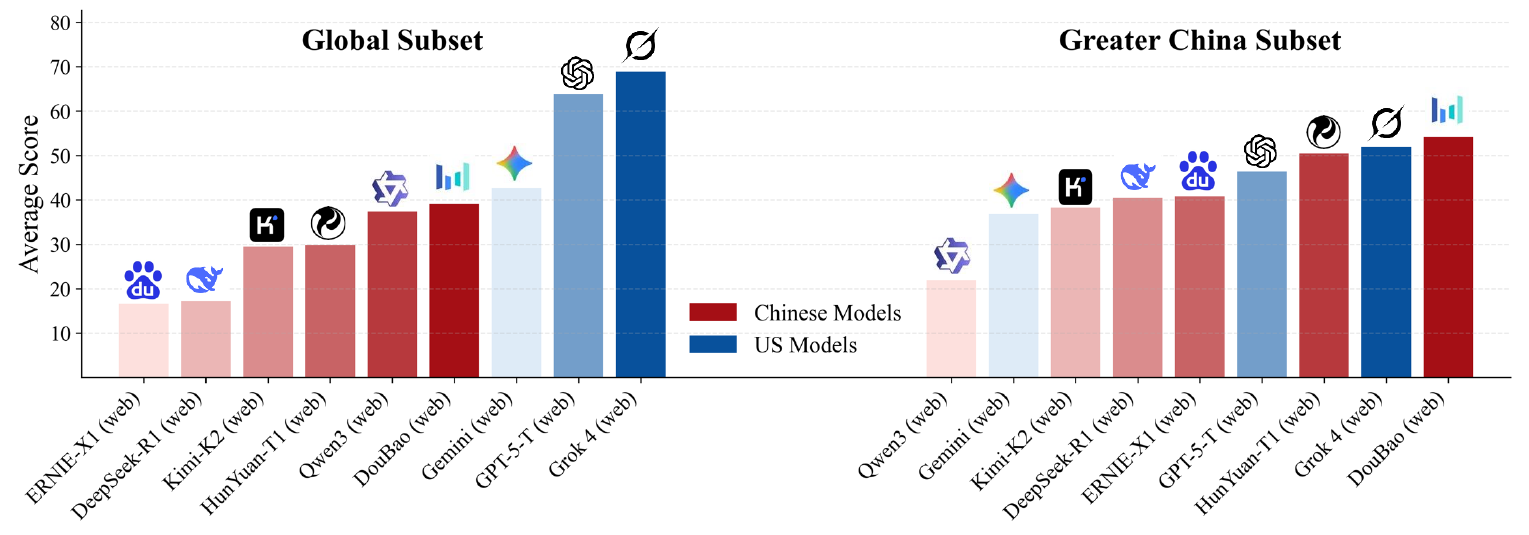}
    \vspace{-0.05in}
    \caption{
    The performance of web-based products on the global subset (left) and the Greater China subset (right) of \benchmark. Note that the performance of human experts is $75.0$ and $88.3$ on the Global and Greater China subsets, respectively.
    }
    \vspace{-0.2in}
    \label{fig:performance}
\end{figure}

\section{Introduction}
\label{sec: intro}
Search is a cornerstone capability for intelligent systems: beyond static knowledge recall, many real-world reasoning problems require agents to acquire, validate, and integrate information from diverse and time-sensitive sources.
Effective decision making hinges on identifying relevant signals, reconciling conflicting evidence, and synthesizing heterogeneous data into a coherent judgment. 
These processes represent the core intellectual skills that large language model (LLM) agents must master---information gathering, coordination, and grounded reasoning.
Yet current benchmarks provide only a partial view of these skills. 
General-purpose browsing benchmarks such as BrowseComp~\cite{wei2025browsecomp} evaluate whether agents can persist through multi-step navigation to uncover obscure facts with short, easily verifiable answers. 
By design, they avoid long-form synthesis and ambiguity resolution and do not assess integration of domain knowledge or multi-source evidence, capabilities that are essential for domain-intensive decision support. 
As a result, they fall short of capturing the multi-step evidence acquisition and reconciliation required in high-stakes, decision-relevant settings.
A natural question, then, is \textit{what environment best reveals whether agents truly possess these skills}.

Finance provides a uniquely demanding answer. Analysts, regulators, and investors routinely engage in searches that combine real-time signals (e.g., market prices, exchange rates) with structured historical disclosures (e.g., annual filings, quarterly reports) and unstructured context (e.g., news events, commentaries). These tasks are time-sensitive, domain-grounded, and decision-relevant: errors in freshness, unit alignment, or source reconciliation directly affect valuations, risk management, and compliance~\cite{nie2024survey-financial,ding2024financial-trading-survey,li-etal-2025-investorbench}. Crucially, financial search is not only a high-value application but also a stress test of general cognitive skills for LLM-based agents. For example, checking ``the latest close price of IBM'' requires rapid retrieval and verification under freshness constraints; retrieving ``Starbucks' total assets as of September 2020'' requires timestamping and accounting alignment; identifying ``the month since 2010 with the largest single-month increase in the S\&P 500'' requires multi-period synthesis, provenance reconciliation, and error-tolerant reasoning. These examples illustrate how financial search combines timeliness, precision, and evidence integration, making it a natural proving ground for assessing whether LLMs can support realistic, high-stakes decision making.

To address this need, we introduce \textbf{\benchmark}, the first \textit{open-domain} benchmark for realistic, analyst-style financial search, comprising $635$ questions that require time-sensitive acquisition and multi-source evidence integration. 
To mirror the day-to-day needs of professional analysts, we design three task families—\textit{Time-Sensitive Data Fetching}, \textit{Simple Historical Lookup}, and \textit{Complex Historical Investigation} (Table~\ref{tab:case}).
These tasks correspond to core analyst workflows. 
\textit{Time-Sensitive Data Fetching} tasks capture situations where the correct response changes rapidly (e.g., stock prices, exchange rates, and gold prices), emphasizing rapid retrieval and verification under tight time constraints. 
\textit{Simple Historical Lookup} tasks reflect frequent point-in-time lookups. For example, an analyst may ask ``What was Apple's iPhone revenue in Q4 2018?'' or ``How many employees did Google report in 2015?'' These questions require going back to the correct historical disclosure and aligning with the company’s reporting calendar.
\textit{Complex Historical Investigation} tasks involve building multi-period views that integrate different sources. For instance, ``Over the past ten years, which quarter showed the fastest growth in Tesla's vehicle deliveries?'' or ``Across 2020–2022, how did Microsoft's cloud revenue trend compared to Amazon's?'' 
Such queries demand stitching together multiple reports, checking consistency across sources, and ensuring that values are comparable across time.

During construction, we engaged $70$ professional financial experts for data annotation, conducted multi-stage verification of prompts and reference answers, and specified detailed, rubric-based scoring guidelines. 
To account for differences in data sources and reporting conventions, \benchmark covers two subsets: Global and Greater China. 
Further, since answers of different LLMs may have different formats and exhibit legitimate minor fluctuations (e.g., revisions or rounding across various sources), we adopt an LLM-based evaluation with rubric-guided judging and explicit tolerance bands, complemented by expert spot checks to ensure the overall correctness.

We evaluate $21$ models (products) on \benchmark, spanning web-enabled products and API endpoints. 
As shown in Figure~\ref{fig:performance}, on the Global subset Grok~4 (web) attains the highest overall score ($68.9\%$), outperforming the runner-up, GPT-5-Thinking (web), by $5.0$ percentage points (pp), yet still trailing human experts by $6.1$~pp. On the Greater China subset, Chinese models perform markedly better: DouBao (web) leads the leaderboard, followed closely by YuanBao-DeepSeek-R1 (web) and Grok~4 (web); nevertheless, all remain more than $34$~pp below human performance. 

Further analyses indicate that equipping agents with web search capabilities and financial plugins improves their performance on \benchmark. A case study further reveals that poor performance often stems from insufficient search depth and the retrieval of outdated information.
Our contributions are as follows:
\begin{enumerate}
    \item We introduce \benchmark, the first fully open-source, end-to-end agent benchmark for \textit{open-domain financial data search}. \benchmark comprises $635$ expert-curated queries spanning global and Greater China markets and three analyst-style task families (Time-Sensitive, Simple Historical, Complex Historical), with multi-stage quality control.
    \item We release a carefully curated benchmark dataset with deterministic gold answers and an fully open-source evaluation harness.
    \item We conduct a comprehensive study of $21$ models (web-enabled products and API endpoints), showing that equipping agents with web search and financial plugins consistently improves performance. Our analysis further identifies recurring failure modes: shallow search, stale or mis-timestamped evidence, cross-unit/currency aggregation, and report-calendar misalignment, offering concrete targets for future improvement. For example, common failures include neglecting to call specialized data plugins in favor of less reliable web searches, extracting incorrect data from a valid source (e.g., confusing opening vs. closing prices), and unnecessarily over-complicating simple queries like ``market cap'' into multiple steps. 
    % \textcolor{blue}{[TODO for XUANLIANG: Give some concrete examples of failure modes]}.
\end{enumerate}

\begin{table*}[t]
    \centering
    \setlength{\tabcolsep}{3pt}
    \small
    \begin{tabular}{lccccc}
\toprule
\multirow{2}{*}{\textbf{Benchmark}} & \textbf{Open-domain} & \textbf{Tool} & \textbf{Time-sensitive} & \textbf{End-to-end}  & \textbf{Holistic} \\
& \textbf{Search} & \textbf{Use} & \textbf{Data} &  \textbf{Agent Evaluation} & \textbf{Evaluation} \\
\midrule
FinQA~\cite{chen-etal-2021-finqa} & \redmark&\redmark &\redmark &\redmark &\redmark \\
ConvFinQA~\cite{chen-etal-2022-convfinqa} & \redmark&\redmark &\redmark &\redmark &\redmark \\
FLUE~\cite{shah-etal-2022-flue} & \redmark&\redmark &\redmark &\redmark &\redmark \\
MultiFinBen~\cite{peng2025multifinben} & \redmark&\redmark &\redmark &\redmark  &\redmark\\
FinanceQA~\cite{mateega2025financeqa} & \redmark&\redmark &\redmark &\redmark  &\redmark\\
BizFinBench~\cite{lu2025bizfinbench} & \redmark&\redmark &\redmark &\redmark  &\redmark\\
FinEval~\cite{guo-etal-2025-fineval} & \redmark & \greencheck &\redmark & \greencheck &\redmark \\
CPA-QKA~\cite{kuang2025FinCDM}  & \redmark & \greencheck &\redmark & \greencheck  &\redmark\\
Finance Agent Benchmark~\cite{bigeard2025financeagentbenchmark} & \greencheck & \greencheck & \redmark & \greencheck  &\redmark \\
\midrule
\textbf{\benchmark} (Ours) & \greencheck & \greencheck & \greencheck & \greencheck  &\greencheck\\
\bottomrule
\end{tabular}
    \caption{
    Comparison of \benchmark with existing financial benchmarks.
    }
    \label{tab:comparison_finbench}
\end{table*}

Taken together, \benchmark enables us to measure, for the first time, how close LLM agents are to expert-level competence in realistic financial search. Our results show that models such as Grok 4 and GPT-5-Thinking can already approach human accuracy in certain subsets, demonstrating the remarkable progress of web-enabled LLMs. At the same time, persistent gaps in freshness awareness, multi-source reconciliation, and temporal reasoning indicate that current systems remain fragile when confronted with the full complexity of analyst-style tasks. In this way, \benchmark not only benchmarks performance in a critical domain, but also highlights the broader aspects of intelligence that today’s LLMs are beginning to approximate---while still falling short of the robustness, adaptability, and judgment required for reliable decision support.

\textbf{Related Works.}\quad 
Recently, many benchmarks are proposed for evaluating browsing capabilities, while they fall short along two axes: 
(\textit{i})~General-purpose browsing benchmarks, like BrowseComp~\cite{wei2025browsecomp}, BrowseComp-ZH~\cite{zhou2025browsecompzh}, and BrowseComp-Plus~\cite{chen2025browsecompplus}, are intentionally domain-agnostic and center on lookup-oriented tasks with short, verifiable targets. 
They optimize for \textit{findability} rather than \textit{analysis}: temporal validity, unit/denomination normalization, reporting-calendar alignment (e.g., TTM vs.\ FY vs.\ quarterly\footnote{Trailing Twelve Months, Fiscal Year, and Quarterly}), and provenance reconciliation across sources are not required, leaving them weakly diagnostic for finance-grade decision support.
(\textit{ii})~Financial QA-style benchmarks (e.g., FinQA~\cite{chen-etal-2021-finqa}, FinanceQA~\cite{mateega2025financeqa}) pre-collect relevant context and bypass open-domain search and tool use, thereby under-assessing agents’ search competence and diverging from analyst workflows \cite{wang2023survey-tat,yang2024finrobot} (see Table~\ref{tab:comparison_finbench}). 
The Finance Agent Benchmark~\cite{bigeard2025financeagentbenchmark} offers an end-to-end evaluation but is confined to a self-constructed system, which is a base model augmented with a retrieval module. We advocate for a holistic evaluation that assesses the performance of web-based products. The unrestricted evaluation, which allows any search tool or source, better reflects the models' practical utility. Moreover, the benchmark's use of only historical data permits success via memorization, not necessarily real-time information retrieval.

\section{\benchmark}
\label{sec:benchmark}
We begin by outlining our design principles and choices. 
We then describe \benchmark’s construction, quality-control measures, and some descriptive statistics.

\subsection{Design Principles}
Before detailing \benchmark, we set out the desiderata for a high-quality financial search benchmark and explain how \benchmark addresses each.
\begin{enumerate}
    \item \textbf{Task professionalism \& diversity.} \quad Financial data retrieval encompasses diverse task types that vary significantly in complexity and time sensitivity. These range from real-time market data queries requiring immediate responses to complex multi-period analytical investigations spanning historical datasets. Given the intricate nature of financial metrics—with nuanced distinctions in reporting standards, calculation methodologies (TTM/FY), and temporal specifications,task design and validation require careful oversight by domain experts to ensure professional accuracy and relevance.
        \begin{itemize}
            \item[$\star$] \textbf{Our design:}\quad  Engaging with various professional financial analysts, we carefully design three tasks that mirror analyst's daily workflow, namely time-sensitive data fetching, simple historical lookup, and complex historical investigation. Details are demonstrated in~\Cref{ssec:benchmark overview}.
        \end{itemize}
    \item \textbf{High quality of questions.} \quad Financial figures vary across sources, vintages, and definitions; ambiguous prompts can admit multiple ``correct'' answers. Without precise definitions, provenance, and reproducible grading criteria (including tolerance for legitimate minor drift due to rounding/revisions), scores reflect dataset noise rather than model ability, harming \textit{reliability} and \textit{fairness}. High-quality items with unambiguous targets and auditable references are therefore prerequisite for credible evaluation.
        \begin{itemize}
            \item[$\star$]  \textbf{Our design:}\quad We perform a set of quality-control processes to ensure the quality of each question, including reliable data source selection, mitigating ambiguity, and multi-expert answer verification. By integrating professional financial expertise. Details are in~\Cref{ssec:quality}.
        \end{itemize}
    \item \textbf{Broad market coverage.} \quad External validity in finance depends on robustness across markets, languages, and regulatory/reporting conventions. Cross-market coverage stresses generalization under heterogeneous tickers, filing formats, calendars/time zones, and currency/denomination regimes, and surfaces failure modes that single-market tests systematically miss. This breadth is essential to assess readiness for real-world deployment rather than a narrow sandbox.
        \begin{itemize}
            \item[$\star$] \textbf{Our design:}\quad \benchmark comprises two subsets, Global (Western markets) and Greater China, with questions in both English and Chinese; see~\Cref{ssec:dataset-statistic} for details. To enable fair cross-market comparison, we mirror task templates across subsets and balance entity coverage by sector and size. Bilingual questions are involved to support cross-lingual evaluation.
        \end{itemize}
\end{enumerate}

\begin{table*}[t]
    \centering
    \small
    \setlength{\tabcolsep}{6pt} % 列间距稍微大一些
    \renewcommand{\arraystretch}{1.25} % 行间距
    \caption{
        Examples of the three tasks in \benchmark, with retrieval depth, temporal span, reasoning complexity, and typical data types.
    }
    \label{tab:case}
    \resizebox{\textwidth}{!}{
   \begin{tabular}{
            >{\raggedright\arraybackslash}m{0.16\textwidth}  % Task
            m{0.34\textwidth}                                % Example
            >{\centering\arraybackslash}m{0.08\textwidth}    % Retrieval Depth
            >{\centering\arraybackslash}m{0.10\textwidth}    % Temporal Span
            >{\centering\arraybackslash}m{0.10\textwidth}    % Reasoning Complexity
            m{0.20\textwidth}                                % Data Types
        }
        \toprule
        \textbf{Task} & \textbf{Example} & \textbf{Retrieval Depth} & \textbf{Temporal Span} & \textbf{Reasoning Complexity} & \textbf{Typical Data Types / Examples} \\
        \midrule
        T1. Time Sensitive Data Fetching 
        & \textit{IBM latest close price. Obtained from a real-time query of IBM} 
        & 1 & 1 day & Easy 
        & Stock prices, FX rates, gold prices (real-time quotes) \\
        
        \rowcolor[gray]{0.98}
        T2. Simple Historical Lookup
        & \textit{What was the total assets of Starbucks as of September 27, 2020? 
        (Answer: \$29374.5 million, rounding errors allowed.)} 
        & 1 & 1 day & Medium 
        & YoY (Year-over-Year), HoH (Half-on-Half), TTM (Trailing Twelve Months), FY (Fiscal Year), quarterly reports \\
        
        \rowcolor[gray]{0.95}
        T3. Complex Historical Investigation 
        & \textit{From Jan 2010 to Apr 2025, in which month did the S\&P 500 index experience the largest single-month increase? 
        (Answer: Apr 2020 (12.68\%), error ±0.1\% allowed.)} 
        & $>$1 & 184 months & Hard 
        & Multi-period views, currency/unit normalization, corporate action adjustments, data provenance \\
        \bottomrule
    \end{tabular}
    }
\end{table*}

\subsection{Task Design of \benchmark}
\label{ssec:benchmark overview}

We define three task types aligned with daily analyst workflows. Each requires at least one external tool call, and each question has a single, fully objective answer. These tasks test core search \& reasoning skills that are not only central to financial analysis but also broadly important to knowledge work in many disciplines, such as journalism, policy research, and scientific data analysis. In all domains, workers must fetch fresh information, verify point-in-time facts, and synthesize evidence across long horizons before drawing conclusions.

\begin{enumerate}[leftmargin=1.2em,label=\textbf{T\arabic*}]
  \item \textbf{Time-Sensitive Data Fetching.}  
  This task type asks for data that changes daily or intraday, such as the latest close, a new filing, or a guidance update. It fits trading, monitoring, and event reactions where decisions depend on the newest number.  
  \emph{Example:} ``What was Nvidia’s closing price yesterday?'' ''Latest change in Dow Jones Industrial index  (based on the closing price of current and previous trading day)''  
  This type stresses freshness management, calendar handling, ticker aliasing, and conflict resolution across sources. Similar challenges arise in real-time journalism, monitoring policy updates, or tracking clinical trial results.

  \item \textbf{Simple Historical Lookup.}  
  This task type asks for a fixed point fact, such as an issuer's FY2024 R\&D expense\footnote{R\&D expense refers to research and development spending, reported in a company’s financial statements. FY2024 means fiscal year 2024.} or TTM revenue\footnote{TTM (Trailing Twelve Months) revenue is the sum of revenue over the most recent 12-month period.} on a given date. It supports baselining, YoY\footnote{YoY (Year-over-Year) compares a financial metric with the same period in the previous year.} or HoH\footnote{HoH (Half-over-Half) compares a metric with the previous half-year period.} comparisons, event studies, and backtests that rely on exact values.  
  \emph{Example:} ``What was Tesla’s reported revenue in Q2 2023?''  
  The key challenges are aligning reporting conventions (FY, TTM, quarterly), handling restatements, and ensuring unit and currency fidelity. Comparable skills are crucial in policy research (e.g., retrieving census data), medicine (e.g., comparing trial endpoints), or academic meta-analysis.

  \item \textbf{Complex Historical Investigation.}  
  This task type asks for multi-period aggregation or synthesis, such as identifying the month with the largest single-month gain for a major index over a long window. It underpins trend analysis, factor research, valuation comps, and risk monitoring.  
  \emph{Example:} ``Over the last 30 years, which month had the steepest decline in the S\&P 500?''  
  The challenges include retrieving across long horizons, adjusting for corporate actions, normalizing units, and composing multi-step reasoning without error. Such synthesis is equally relevant in climate science (long-horizon weather anomalies), history (identifying peak conflict years), or epidemiology (largest single-month case surges).
\end{enumerate}

Detailed comparison is shown in~\Cref{tab:case}.
Together these task types cover three critical capabilities: freshness management, point-in-time fidelity, and multi-period synthesis. They mirror the actual workflows of analysts, are grounded in real data and conventions, and scale in difficulty from T1 to T3. This progression enables fine-grained error analysis and highlights how benchmarking these skills matters not just for finance, but for intelligence in knowledge work broadly.

\textbf{Time Cost of Financial Analysts on These Three Tasks.}\quad 
Financial analysts worldwide dedicate substantial resources to these core information retrieval activities. There are approximately 370,000 financial professionals in the US (based on Bureau of Labor Statistics) and probably over 1 million globally---including equity researchers, portfolio managers, risk analysts, and investment bankers---who regularly perform these tasks as part of their daily workflow.

For T1 (Time-Sensitive Data Fetching), analysts typically spend around 1-2 minutes per query for quick references. 
This task is also commonly performed by non-expert investors in scenarios such as checking the current price of their investments.

For T2 (Simple Historical Lookup) represents the most frequent task type, with individual analysts performing 10-30 such queries daily for financial analysis, peer comparisons, and modeling. Each lookup averages 5-10 minutes, accounting for data validation (faster for standard financials that can be retrieved from filings, slower for less common financial data such as operational, macroeconomic, and industry data).

For T3 (Complex Historical Investigation) demands the highest time investment per-query, often requiring 15-60 minutes for comprehensive data retrieval and calculation, based on the complexity of the data retrieval and calculation steps. While less frequent (fewer than 20 queries per analyst monthly), these investigations and calculations are important building blocks in financial analysis and reports.

While standardized templates and automated tools already exist to facilitate these tasks---such as comparable company analysis frameworks that can be efficiently updated---approximately half of these information retrieval activities still inevitably require manual data collection and custom analytical framework development. 
If AI models could accurately perform such tasks, analysts could further automate these processes and significantly enhance overall productivity.

\begin{figure}[t]
    \centering
    \includegraphics[width=0.85\linewidth]{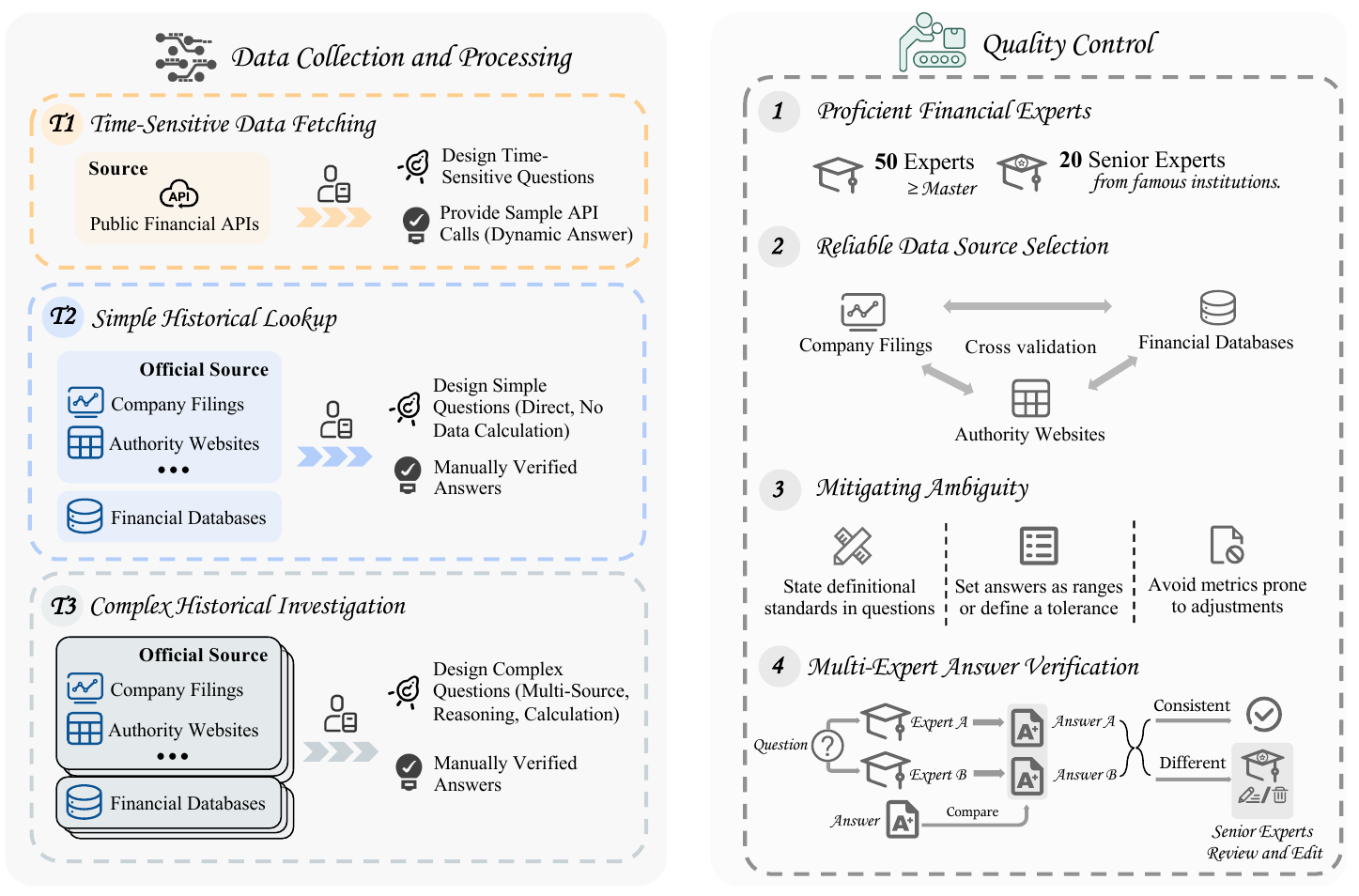}
    \caption{
    The overview of the construction process.
    % \benchmark的构建过程
    The construction of this benchmark involves three distinct tasks. 
    % 其中3类数据采用了不同的源数据和处理方法，以及都采用了一致的质量控制
    The data for each task originate from different sources and undergo separate processing pipelines. 
    A uniform quality control procedure is applied across all tasks.
    }
    \label{fig:pipeline}
\end{figure}

\subsection{Construction of \benchmark}
\label{ssec:data}

% \subsubsection{Source Collection}
% \begin{description}[leftmargin=1.8em]
%     \item[\textit{Market feeds}] {TODO: For T1 time-sensitive data, questions are handcrafted and results are based on yfinance API, which we have checked real-time accuracy against other professional financial databases. We choose yfinance because its is open source and accurate for US assets}
%     \item[\textit{Listed Company filings}] {US listed company filies such as 10-K, Chinese listed company filings, Chinese~CSRC, \dots}
%     \item[\textit{Authorities websites}] {Federal Reserve Bank, The World Bank, Organisation for Economic Co-operation and Development (OECD),US Census Bureau etc.}
%     \item[\textit{Professional financial database}] {Historical market data and data sheets used in construction of T3 Complex historical analysis are generally collected from profession financial database to ensure accuracy.}

% \end{description}

% \subsubsection{Question and Answer Generation}
To accommodate the unique characteristics of different tasks, we employ a variety of question-and-answer construction strategies to ensure both diversity and quality, as shown in Figure~\ref{fig:pipeline} (left side).  
% We also asked a human financial expert to solve T2 and T3, and they achived an average success rate of 81\% and 64\% upon first attempt, respectively. 
% 下面我们分别介绍3个子任务的数据收集以及处理过程
We now detail the data collection and processing procedures for the three tasks.

\textbf{For Time-Sensitive Data Fetching.}\quad
Time sensitive data includes \textit{real-time} stock prices, index levels, exchange rate, metal prices etc. Some of the example questions include:

\begin{itemize}
    \item ``Latest closing price of Bloom Energy(NYSE)''
    \item ``The latest opening price of Starbucks''
    \item ``The latest percentage change of Rivian (NASDAQ), based on the latest closing price and the previous closing price''
    \item ``USD/THB price today''
    \item ``Latest price of S\&P 500''
\end{itemize}

Financial experts first mannually design questions asking for time-sensitive data that can be verified through API. Time-sensitive data changes over time, so we actually prepare the code for API calls that obtain real-time data points for each question. Finally, financial experts check each API result against the real-time data to ensure the retrieved result is correct and on-time, for accurate evaluation. We establish permissible error margins for evaluation based on the specific volatility characteristics of different assets (e.g., equities, forex)

\textbf{For Simple Historical Lookup.}\quad
Simple historical data includes historical market data (stock price, oil price etc.), corporate financials, macro economic statistics that can be directly obtained from official sources \textit{without data processing}. 
Some of the example questions include:

\begin{itemize}
    \item ``What was the additional paid-in capital of Lands' End as of the end of the fiscal year 2020? (answer in thousand dollars, rounded to the nearest integer)''
    \item ``What was the closing value of the VIX on April 25, 2022? (rounded to two decimal places)''
    \item ``For the fiscal year 2023, what was Planet Labs' Net cash provided by investing activities? (please answer in thousands of dollars, rounded to the nearest integer)''
    \item ``What was the U.S. Housing Market Index (HMI) in November 2014? (answer rounded to the nearest integer)''
    \item ``In April 2015, what was the CPI of Russia? (base year 2015=100, rounded to two decimal places)''
\end{itemize}

We collect data for this task from two sources to enrich the diversity. (i) Financial experts select documents from official sources (e.g., listed company filings, regulatory authority websites, statistics bureau) and extract key data points to formulate questions and their corresponding answers. (ii) We also design questions using historical data with consistent definitions from professional financial databases. To mitigate the risk of data revisions for certain official statistics (e.g., macroeconomic indicators), we address potential ambiguity by setting a reasonable answer range or by explicitly specifying the reference time point in the question.

\textbf{For Complex Historical Investigation.}\quad 
Complex historical data includes financial data that needs to be derived based on multiple historical financial data points, and requires calculation and reasoning to solve. Some of the example questions include:

\begin{itemize}
    \item ``What were the specific dates from January 1, 2020, to December 31, 2024, when London Gold (XAUUSD) dropped by more than \$80 in a single day? Please list these dates and the corresponding daily drop in USD (rounded to the nearest integer), presented in a table sorted by date in ascending order.''
    
    \item ``During April 2025, did the daily changes (compared to the previous day) in the central parity rates of EUR/CNY, HKD/CNY, and USD/CNY always occur in the same direction (i.e., all rates increasing together or all rates decreasing together)? Among the days when these rates did not move in unison, find the exact date when the USD/CNY central parity rate experienced its greatest single increase. List the three central parity rates (EUR/CNY, HKD/CNY, USD/CNY) for that specific day, rounded to four decimal places.''
    
    \item ``Which constituent stock of the Nasdaq 100 Index (NDX), with a Price-to-Earnings Trailing Twelve Months (PE-TTM) greater than 0 and less than 20 as of the market close on September 30, 2024, reported the largest operating revenue (consolidated financial statements, in billions of USD) in its 2024 interim report? Please provide the stock name and its operating revenue. Unit: billion USD, rounded to two decimal places.''
    
    \item ``What is the year-over-year change in the proportion of Johnson \& Johnson's revenue from international markets (excluding the United States) for each of the past three years (2022–2024)? Please provide the change in percentage points, rounded to two decimal places.''
\end{itemize}

The construction process involves two primary methods. (i) Financial experts design questions based on their real-world professional scenarios and get answers using reliable financial data sources. (ii) Financial experts download tables from a verified and reliable financial database and annotate questions based on the tables. Furthermore, financial experts screen and refine these candidates, finalizing 2 to 5 high-quality questions on each table and recording their definitive answers.

\subsection{Quality Control}
\label{ssec:quality}
To ensure \benchmark's quality, we implemented a rigorous quality-control process throughout construction and summarize the key measures below.

\textbf{Proficient Financial Experts.}\quad 
Our 70-person expert cohort is comprised of a 50-expert annotation panel and a 20-expert senior review panel. We assembled the former as a distinguished group of 50 financial experts to conduct benchmark annotations. All panel members possess advanced degrees in finance (minimum Master's level) and maintain active professional standing within the financial services industry. Panel selection follows a stringent qualification protocol, whereby candidates undergo comprehensive domain-specific assessments to ensure annotation quality and inter-annotator reliability. The latter, our senior review panel, consists of 20 senior financial experts who handle discrepancies arising from blind review. The entire expert cohort includes practitioners from prestigious institutions such as Citadel, J.P. Morgan, Deutsche Bank, Nomura, Citigroup, Credit Suisse, Harvest Fund, Zhongou Asset Management, CITIC Securities, and Huatai Securities.

\textbf{Reliable Data Source Selection.}\quad 
To ensure the authority and accuracy of our data, all the answers are sourced from highly reliable channels, including official filings of listed companies, government and regulatory authority websites, and professional financial databases. 
We employ a multi-source cross-validation method to ensure data reliability and eliminate ambiguity. 
For instance, we cross-reference data from two different official websites or validate data from a professional financial database against an official website. 
Through this process, we identify that even some widely recognized professional financial databases contain inconsistencies in definitional standards or calculation errors and avoid them. 

\textbf{Mitigating Ambiguity.}\quad 
To address ambiguities arising from inconsistent calculation methods for the same metric across different institutions, we avoid questions prone to such variation. 
For instance, the methodologies for calculating forward-adjusted stock prices can differ significantly across data providers, and the precise definition of 'Earnings' in Price-to-Earnings (PE) Trailing Twelve Months (TTM) ratios often varies. 
A more comprehensive list of such cases is detailed in Appendix~\ref{subapp:Inconsistent Calculation Methods}.

Furthermore, to ensure the unambiguity of our questions, we implement the following measures. We also summarize the detailed guide in Table~\ref{tab:ambiguity}.
(\textit{i}) We explicitly state definitional standards within the question itself (e.g., specifying Static PE vs. PE TTM, or Nominal GDP vs. Real GDP) and avoid metrics with ambiguous time-points, such as the prices of assets with 24-hour trading cycles.
(\textit{ii}) We set answers as numerical ranges or define a tolerance for precision to accommodate minor discrepancies that may arise from different calculation tools or rounding methods.
(\textit{iii}) We mitigate risks from data revisions by avoiding, where possible, metrics prone to retrospective adjustments (e.g., GDP, Non-Farm Payrolls). If their inclusion is necessary, we formulate the answer as a range.

\textbf{Multi-Expert Answer Verification.}\quad 
% We have established a rigorous answer verification process to ensure the accuracy of both questions and answers. 
The answer verification mechanism utilizes a blind review module.
% First, a financial expert designs a question and its standard answer. 
After obtaining a question and its answer, one or two other financial experts solve the question independently without access to the answer key. 
If discrepancies arise in the results or if an expert deems a question to be ambiguous, a senior expert arbitrates the matter. 
Based on the final judgment, the question or answer will be modified, or the question will be discarded entirely. 

The dataset construction process encompassed approximately 180 hours of contributions from financial experts and 60 hours from senior financial experts, thereby ensuring comprehensive professional oversight throughout development.

\subsection{Data of \benchmark}
\label{ssec:dataset-statistic}
%  \begin{wraptable}{R}{0.45\textwidth}
%     \centering
%     \small
%     % \vspace{-4mm}
%     \input{table/statistics}
%     \caption{Data statistics of \benchmark.} 
%     % \vspace{-1em}
%     \label{tab:statistics}
% \end{wraptable}

\begin{figure}[t]
    \centering % 将整个 figure 环境居中
    \begin{subfigure}[b]{0.48\textwidth}
        \centering
        \includegraphics[width=\linewidth]{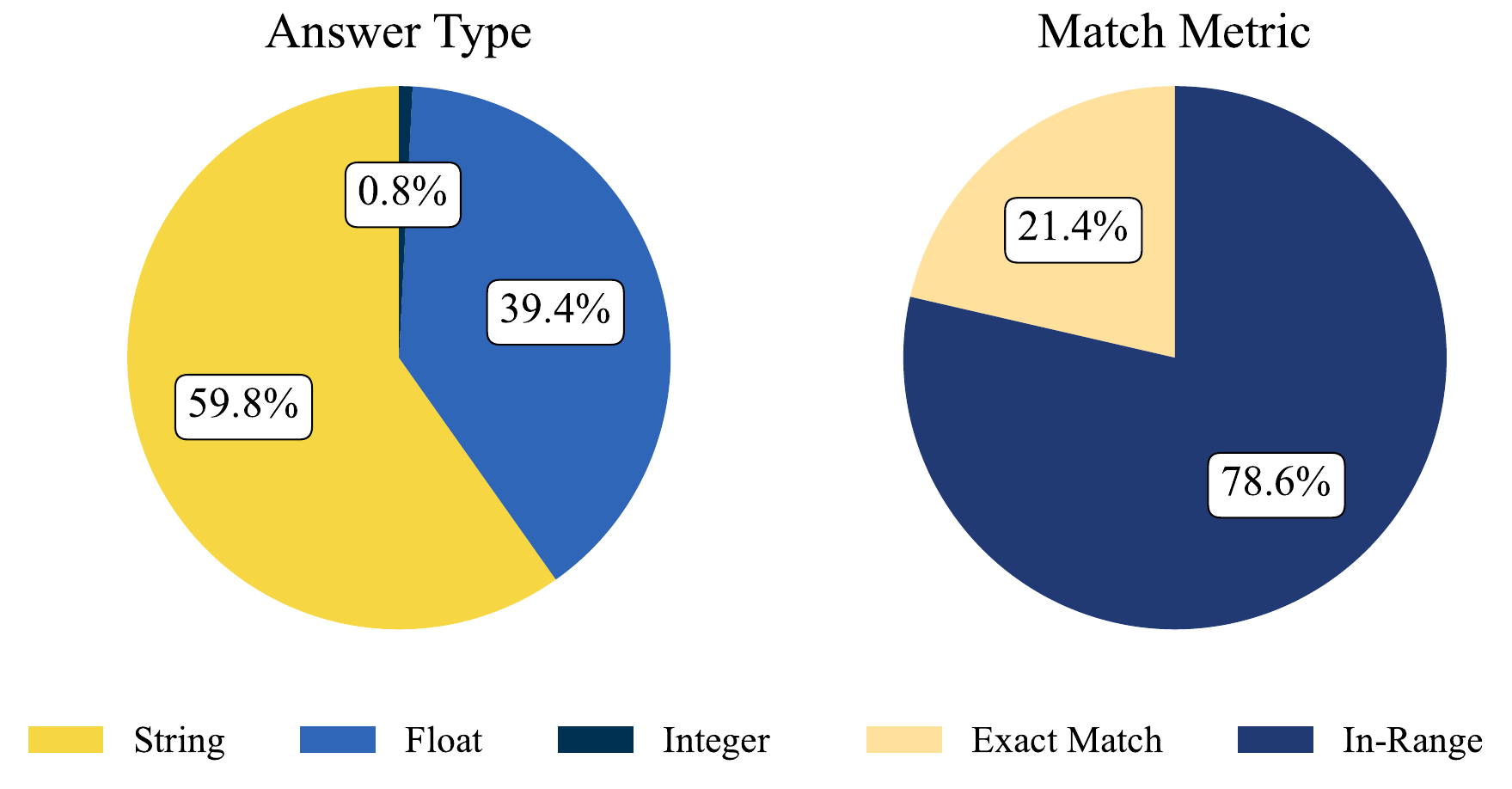}
        \caption{Distribution of the answer type and evaluation rubrics.}
        \label{fig:metrics}
    \end{subfigure}
    \hfill 
    \begin{subfigure}[b]{0.48\textwidth}
        \centering
        \includegraphics[width=\linewidth]{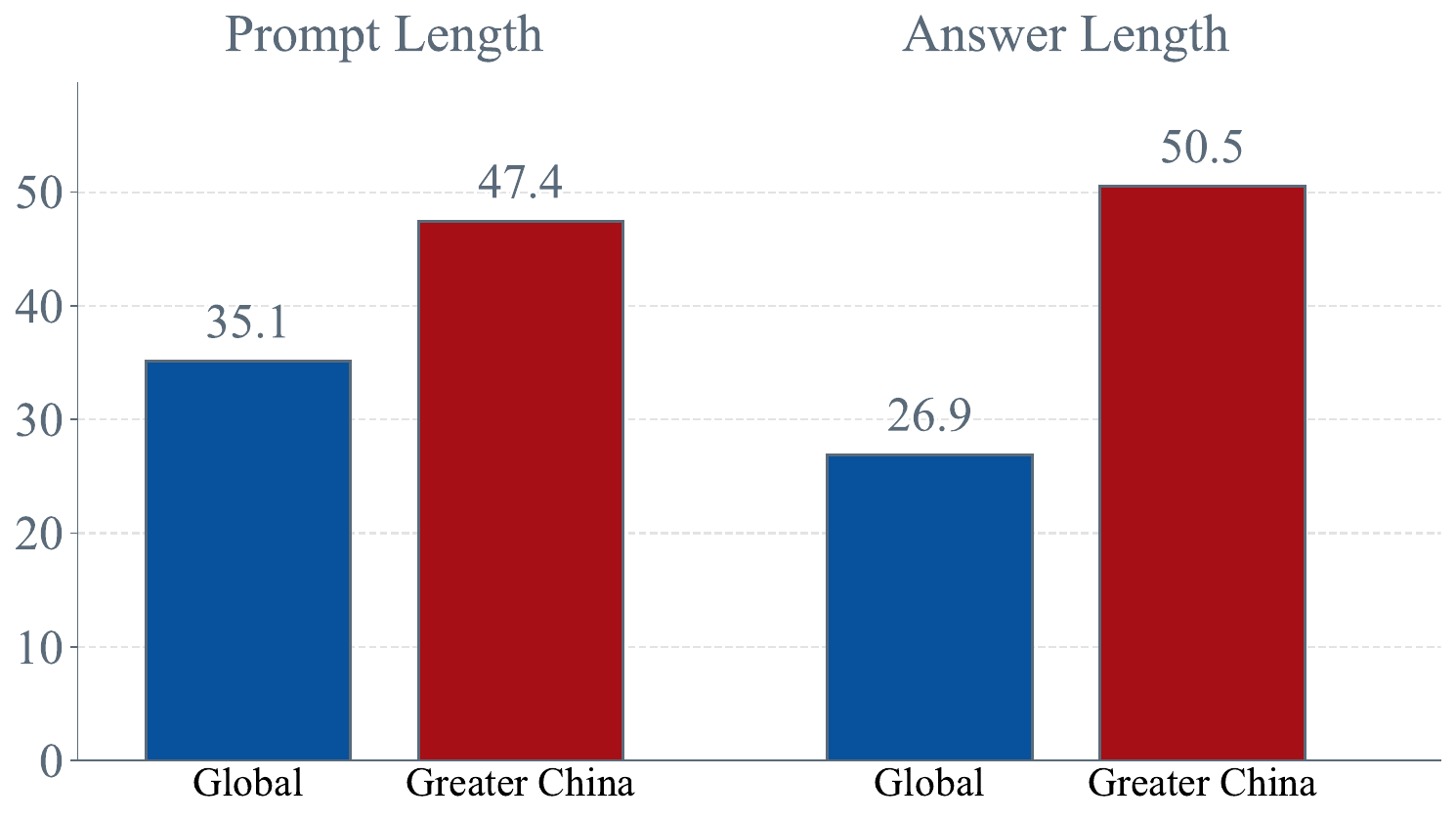}
        \caption{Average length of prompts and answers on each subset.}
        \label{fig:answer_len}
    \end{subfigure}
    \caption{Data statistics of \benchmark.}
    \label{fig:statistics}
\end{figure}

\begin{figure}[t]
    \centering
    \begin{subfigure}[b]{0.45\textwidth}
        \centering
        \includegraphics[width=\linewidth]{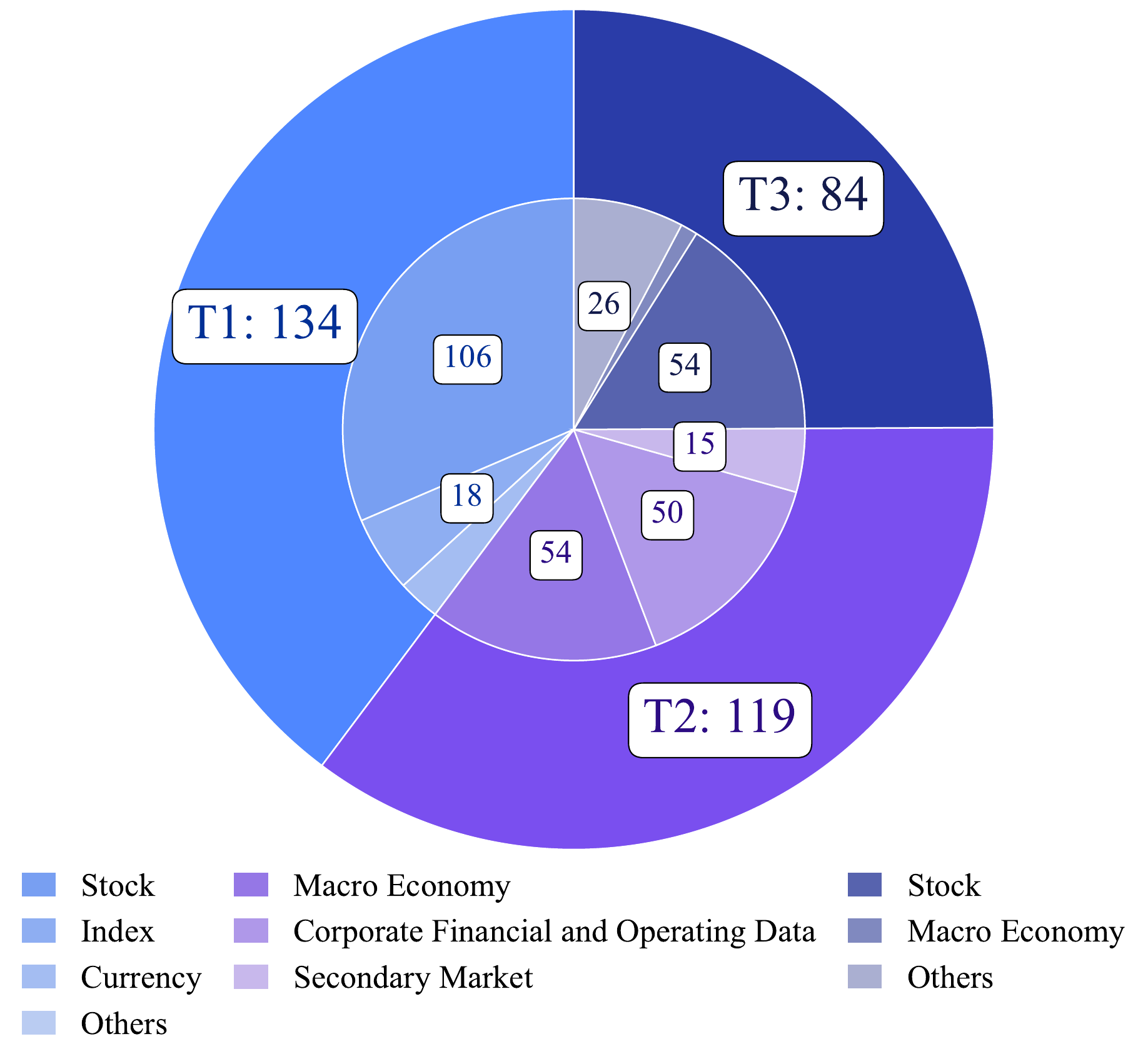}
        \caption{Topic distribution of Global subset.}
        \label{fig:tag_distribution_global}
    \end{subfigure}
    % \hfill 
    \begin{subfigure}[b]{0.45\textwidth}
        \centering
        \includegraphics[width=\linewidth]{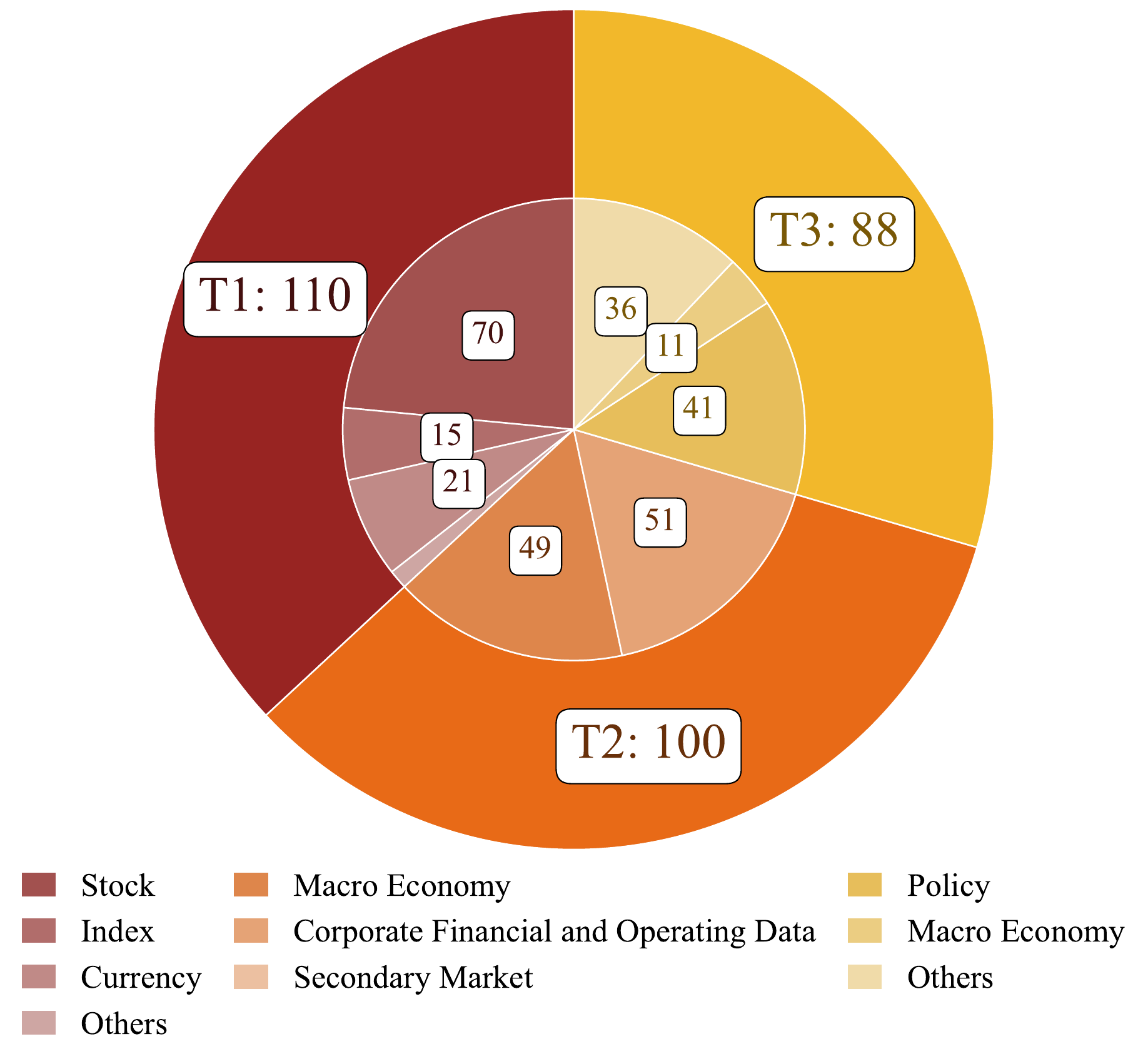}
        \caption{Topic distribution of Greater China subset.}
        \label{fig:tag_distribution_domestic}
    \end{subfigure}
    \caption{Topic distributions in \benchmark.}
    \label{fig:tag_distribution}
\end{figure}

% 我们在这一节展示我们benchmark上的数据统计
In this subsection, we present the statistics of \benchmark. 
% 我们在表1中展示了数据集中基本的数据信息
We summarize the basic statistics in Figure~\ref{fig:statistics}. 
% 可以发现，\benchmark中大部分评测标准允许数值在一定范围内波动，范围为问题specfic的，由专家手动标注
Most evaluation metrics in \benchmark rely on problem-specific, expert-annotated ranges of acceptable values, rather than a single ground truth.
% 并且，我们在图1中展示了数据集中topic的分布，展示了我们数据集覆盖15个话题的多样性
Furthermore, we illustrate the distribution of topics in Figure~\ref{fig:tag_distribution}, where experts annotated the topic of each question.
The distribution demonstrates the diversity of \benchmark, which covers $10$ distinct topics.

\subsection{Evaluation of \benchmark}
% \textcolor{blue}{[TODO for KAIYUAN: Add our backtesting results here to show that our metric is accurate.]}

\textbf{Evaluation Protocol.}\quad 
Considering the dynamic nature of answers and the need for numerical tolerance in \benchmark, we adopt LLM-as-a-Judge~\cite{zheng2023judging} for evaluation. 
The specific evaluation methods for the three tasks are detailed below, with the corresponding prompts provided in the Appendix~\ref{appendix:prompt}.

For \textit{Time-Sensitive Data Fetching}, we address several challenges: (\textit{i}) time lags between the model response and the evaluation,  (\textit{ii}) potential data latency from some financial APIs, and  (\textit{iii}) the inability of most APIs to query prices at a specific second. To mitigate these time-sensitivity issues, we initiate the evaluation process uniformly after the relevant markets have closed.
To ensure fairness and accuracy, we establish differentiated evaluation rubrics based on the characteristics of various asset classes:

\begin{itemize}
    \item Mainstream Market Stocks and Indices (e.g., U.S., A-shares, H-shares): Evaluation is conducted during non-trading hours. Only minor discrepancies attributable to rounding are permitted.
    \item Other Regional Indices: An answer is considered correct if its value falls within the day's high-low price range.
    \item Foreign Exchange (FX) Rates: To account for potential discrepancies across different data providers, the valid range for an answer is defined as the high-low range of the day, augmented by an additional buffer.
\end{itemize}

For \textit{Simple Historical Lookup} and \textit{Complex Historical Investigation}, which feature static and deterministic answers, we also annotate rubrics, such as a predefined error margin.

\textbf{Evaluation Metrics.}\quad 
% \textcolor{blue}{Add some equations for the metric.}
In \benchmark, we employ the 0-1 error as the metric. 
For questions in Time-Sensitive Data Fetching, we first obtain the real-time answer by executing the API commands.
For questions in other tasks, the answers are static.
After getting the answer, we adopt the LLM to judge referring to the rubrics.
The LLM's assessment is modeled as a judgment function, $\mathcal{J}$, which maps a candidate answer and a set of rubrics to a binary outcome. Let $\mathcal{A}$ be the space of all possible answers and $\mathcal{R}$ be the space of all possible rubric sets. The judgment function $\mathcal{J}: \mathcal{A} \times \mathcal{R} \to \{1, 0\}$ returns 1 if answer $A \in \mathcal{A}$ satisfies the criteria specified in rubrics $R \in \mathcal{R}$, and 0 otherwise.
The final evaluation score, $S$, is derived directly from the output of this judgment function. We define the score using the indicator function $\mathbb{I}(\cdot)$, which formally connects the LLM's logical evaluation to a numerical score. The score is therefore defined as follows:
\begin{equation}
S(A, R) = \mathcal{J}(A, R).
\end{equation}

\begin{table}[htbp]
\centering
\small
\begin{tabular}{lcc}
\toprule
Task & Subset & Accuracy (\%) \\
\midrule
Time-Sensitive Retrieval & Overseas & 91.5 \\
                         & Domestic & 91.7 \\
Simple Historical Lookup & Overseas & 96.8 \\
                         & Domestic & 95.5 \\
Complex Historical Investigation & Overseas & 97.4 \\
                                 & Domestic & 99.8 \\
\bottomrule
\end{tabular}
\caption{Accuracy of LLM-as-a-Judge compared with human evaluation.}
\label{tab:judge-accuracy}
\end{table}

\textbf{Evaluation Accuracy.}\quad  
To validate the reliability of LLM-as-a-Judge, we conducted a human evaluation on a representative subset.  
For each benchmark dataset, we selected 4--5 models and verified their complete evaluation sets, resulting in roughly 400 instances per dataset.  
On this sample, the judgments of LLM-as-a-Judge reached 95\% agreement with human-verified labels, confirming the robustness of our evaluation protocol.  
Detailed results are summarized in Table~\ref{tab:judge-accuracy}.
\section{Experiments}
In this section, we present the main results for \benchmark.
% We evaluate the performance of $22$ mainstream models on \benchmark from August $1^{st}$ to $20^{th}$, which are categorized into two groups:
We evaluate the performance of $22$ mainstream models (products) on \benchmark from August $1^{st}$ to $20^{th}$, and to ensure temporal comparability across models, all T1 evaluations were conducted after the official market close (local market time) on each evaluation day; these models are categorized into two groups:

\begin{itemize}
    \item \textbf{Web-based products (12 models):} Grok 4 (web)~\cite{grok4}, GPT-5-Thinking (web)\footnote{abbr. GPT-5-T (web)}, Gemini-2.5-pro (web)~\cite{google_gemini25pro0506}\footnote{abbr. Gemini (web)}, Qwen3-235B-A22B-2507 (web)\footnote{abbr. Qwen3 (web)}~\cite{qwen3_235b}, DeepSeek-R1 (web)~\cite{deepseek_r1}, DouBao (web)~\cite{doubao}, DouBao-Thinking (web)\footnote{abbr. DouBao-T (web)}, YuanBao-HunYuan-T1-Thinking (web)\footnote{abbr. HunYuan-T1 (web)}~\cite{yuanbao}, YuanBao-DeepSeek-V3 (web)\footnote{abbr. YuanBao-V3 (web)}, YuanBao-DeepSeek-R1 (web)\footnote{abbr. YuanBao-R1 (web)}, Ernie-X1 (web)~\cite{ernie}, and Kimi k2 (web)~\cite{kimi}.
    
    \item \textbf{APIs (9 models):} Gemini2.5-pro (API)\footnote{abbr. Gemini (API)}, Gemini-2.5-pro (Google Search) (API)\footnote{abbr. Gemini-G (API)}, DouBao (API), DouBao-Thinking (API)\footnote{abbr. DouBao-T (API)}, Qwen3-235B-A22B-2507 (API)\footnote{abbr. Qwen3 (API)}, DeepSeek-R1 (API), HunYuan-T1 (API), Ernie-X1 (API), and Kimi k2 (API).
\end{itemize}

% 我们雇佣了不同于标注数据的50名金融专家，令他们在配备搜索工具条件下完成\benchmark中的问题
To establish a human performance baseline, we engage another $50$ financial experts who are not involved in the data construction. 
These experts perform the benchmark tasks while utilizing search tools. 
Their average score is taken as the human baseline.

\subsection{Main Results}
We report the overall performance of the strongest model from each provider in~\Cref{fig:performance}, with the whole performance in Appendix~\ref{appendix: Detailed Scores on benchmark}. The models form a clear performance hierarchy, with a sizable gap to human experts remaining. Rankings differ between the global and Greater China subsets, likely reflecting differences in training-corpus coverage, language/domain alignment, and retrieval infrastructure. On the global subset, \textit{Grok-4 (web)} and \textit{GPT-5-Thinking} form a clear leading tier, with Grok-4 (web) securing the top score and approaching expert-level accuracy. While Gemini-2.5-pro (web) decline when moving from the global to the Greater China subset, \textit{Grok 4 (web)} remains competitive. On the Greater China subset, \textit{DouBao (web)} and \textit{YuanBao-HunYuan-T1-Thinking (web)} are strong on the Greater China subset, though they still trail human experts by a substantial margin. Detailed results and analyses are shown in~\Cref{ssec:result-different-task} and~\Cref{sec:discussion}.

% We first compare the performance of different models, the underlying reasons for which are explored in \S\ref{sec:discussion}.
% Overall, Grok 4 achieves the highest average score ($60.4$), demonstrating remarkable financial search strength. Conversely, ERNIE-X1 registers the lowest overall average ($28.7$).

% In the Global subset, performance is led by Grok 4, which obtains the highest average score of $68.9$, notably achieving the peak performance in T3 with a score of $51.2$. 
% % GPT-5 and Gemini2.5-pro also demonstrate strong performance, securing the highest score in T1 ($66.4$). However, its efficacy dramatically decreases in T3 ($11.9$), contributing to a lower global average compared to Gemini2.5-pro.
% % The lowest performance in the Global subset is observed from ERNIE-X1 and DeepSeek-R1, with average scores of $16.6$ and $17.2$, respectively. DeepSeek-R1, in particular, recorded the lowest score in T1 ($17.9$), suggesting a significant weakness in this specific global task.
% The Greater China subset showcases a different performance hierarchy. DouBao (Non-Thinking) is the undisputed top performer with an average of $54.2$, achieving the highest score in T1 ($88.3$). Similarly, YuanBao (DeepSeek-R1) excells, attaining the second-highest domestic average ($52.5$). 

\subsection{Results Across Different Tasks}
\label{ssec:result-different-task}
\begin{figure}[t]
    \centering
    \begin{subfigure}[b]{\linewidth}
        \centering
        \includegraphics[width=\linewidth]{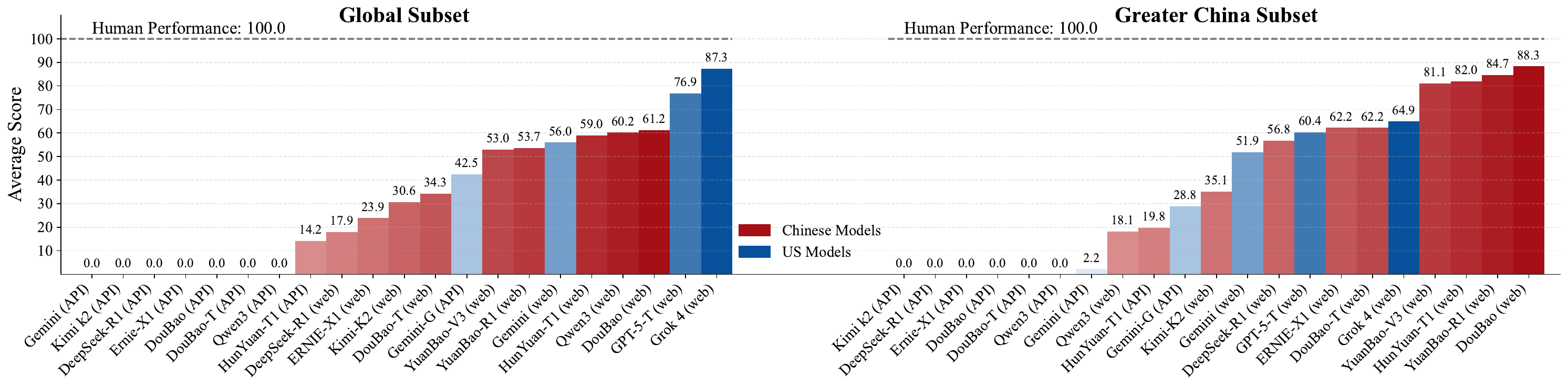} 
        \caption{ Task 1. Time-sensitive data fetching. 
        }
        \label{fig:performance_T1}
    \end{subfigure}
    \begin{subfigure}[b]{\linewidth}
        \centering
        \includegraphics[width=\linewidth]{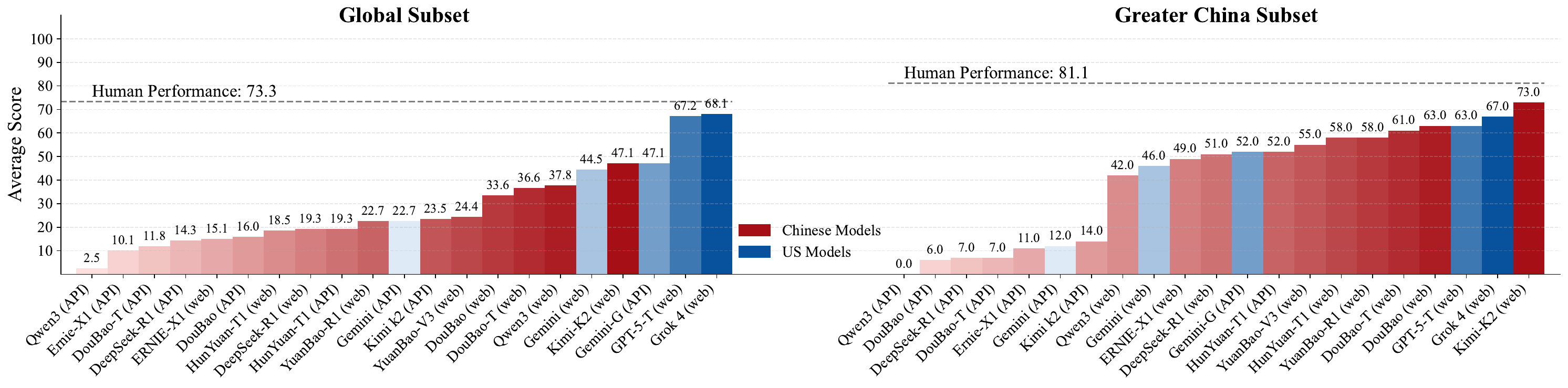}
        \caption{
        Task 2. Simple historical data lookup.
        }
        \label{fig:performance_T2}
    \end{subfigure}
    \begin{subfigure}[b]{\linewidth}
        \centering
        \includegraphics[width=\linewidth]{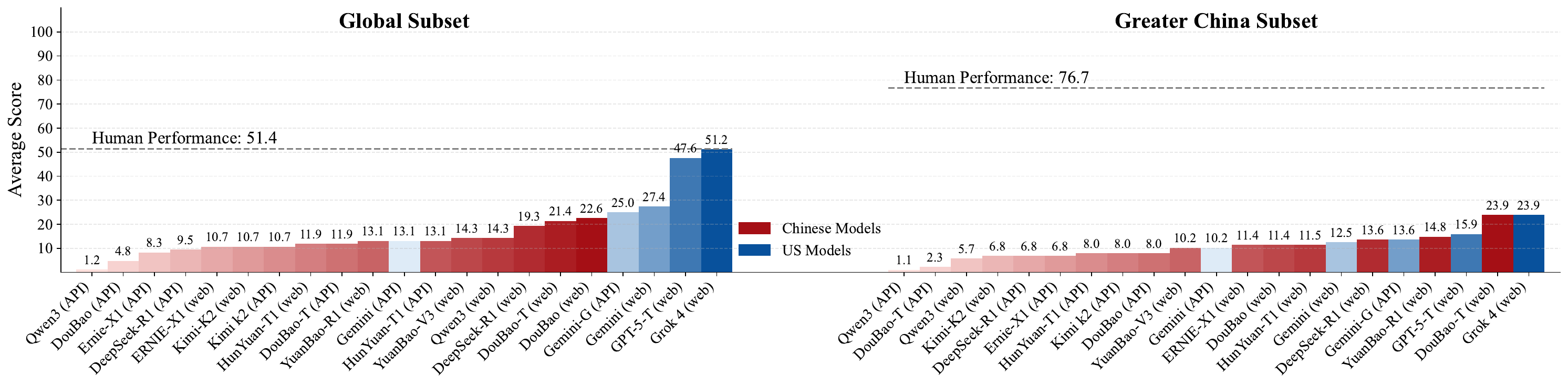}
        \caption{
        Task 3. Complex historical investigation
        }
        \label{fig:performance_T3}
    \end{subfigure}
    
    \caption{
    The performance of various models across the three tasks on \benchmark. Models with $0$ scores are all APIs.
    }
    \label{fig:main}
\end{figure}

We show the results of each tasks respectively in~\Cref{fig:main}.
Our main findings are as follows.

\textbf{Finding 1. Task difficulty increases from T1 to T3.}\quad 
Across models, performance declines \textit{monotonically} from T1 (time-sensitive data fetching) to T2 (simple historical lookup) to T3 (complex historical investigation), indicating that our task design mirrors the escalating demands of professional financial-analysis workflows. This pattern demonstrates that we are probing \textit{complex search and reasoning}: T3 requires multi-hop retrieval across heterogeneous sources and time periods; temporal reasoning (event dating, fiscal–calendar alignment, handling revisions/restatements); fine-grained entity resolution (issuer/ticker/subsidiary/renamed entities); and reconciliation of partial or conflicting evidence, forcing systems to plan, verify, and synthesize rather than merely retrieve. Moreover, success on T2–T3 hinges on finance-specific expertise, including interpreting primary filings and disclosures (10-K/10-Q/8-K), earnings releases and footnotes, distinguishing GAAP vs.\ non-GAAP metrics, and understanding corporate actions (splits, spin-offs, mergers), without which methods commonly fail via stale or misaligned time windows, misread accounting terminology, or incorrect consolidation across corporate structures.

\textbf{Finding 2. US models lead on the global set; Chinese models lead on the Greater China subset.}\quad We attribute this pattern mainly to corpus geography (English/SEC/multinational coverage vs.\ CN/HK/TW disclosures and regulator texts), linguistic and market conventions (domain terminology, tokenization, date and identifier formats) that ease in-region entity resolution, and alignment/recency effects, collectively boosting home-field performance without implying leakage.

% \textbf{Finding 3. Despite broad underperformance versus human experts, Grok-4 approaches expert-level results on the global subset.}

\textbf{Finding 3. Despite broad underperformance versus human experts, Grok-4 and GPT-5-Thinking approaches expert-level results on the global subset.}\quad On the global subset,The outperformance of \textit{Grok-4 (web)} and \textit{GPT-5-Thinking (web)} over other systems becomes more pronounced on more difficult tasks (T1 $\rightarrow$ T2 $\rightarrow$ T3), with its largest margin on T3, indicating it goes beyond retrieval by performing multi-step reasoning, aligning timelines (event dating and fiscal/calendar consistency), and carefully disambiguating entities; on the Greater China subset, while aggregate accuracy still trails experts, Grok 4 (web) attains the \textit{top} score on the most difficult task (T3), reinforcing that its gains reflect genuine reasoning strength rather than surface-level search.

\section{Case Study}
\label{sec:discussion}

In this section, we conduct case studies to analyze the performances in detail.

\subsection{How much do search capabilities impact performance on \benchmark?}

\begin{figure}[t]
    \centering
    \includegraphics[width=0.85\linewidth]{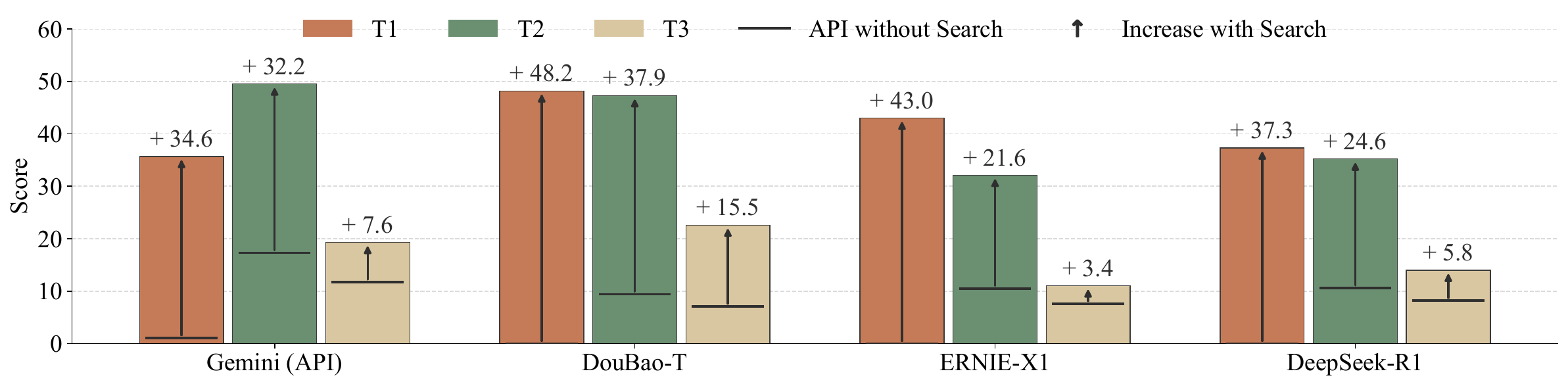}
    \caption{
    % 具备搜索能力的模型相比不具备搜索能力的模型的性能提升
    Performance improvement of search-augmented models over models without search capabilities.
    }
    \label{fig:search}
\end{figure}

As shown in~\Cref{fig:search}, models without search uniformly score $0$ on T1, as they cannot retrieve current financial data. Without search, they still obtain non-zero but low scores on T2 and T3; we attribute this to parametric memory from pre-training (e.g., annual reports and statistical-agency releases), which surfaces approximate facts that are often outdated or misaligned, yielding higher error rates. With search enabled, average gains of $40.8$, $29.0$, and $8.1$ points are observed on T1, T2, and T3, respectively—largest for time-sensitive tasks but still material for complex historical investigations. 

These patterns indicate that \benchmark stresses complex search and reasoning, where success requires planning multi-step queries, aligning timelines and identifiers across sources, and resolving conflicting evidence. In turn, performance reflects not only access to documents but also the ability to verify, synthesize, and reason.

\begin{figure}[t]
  \centering
  \begin{minipage}[b]{0.45\textwidth}
    \centering
    \includegraphics[width=\linewidth]{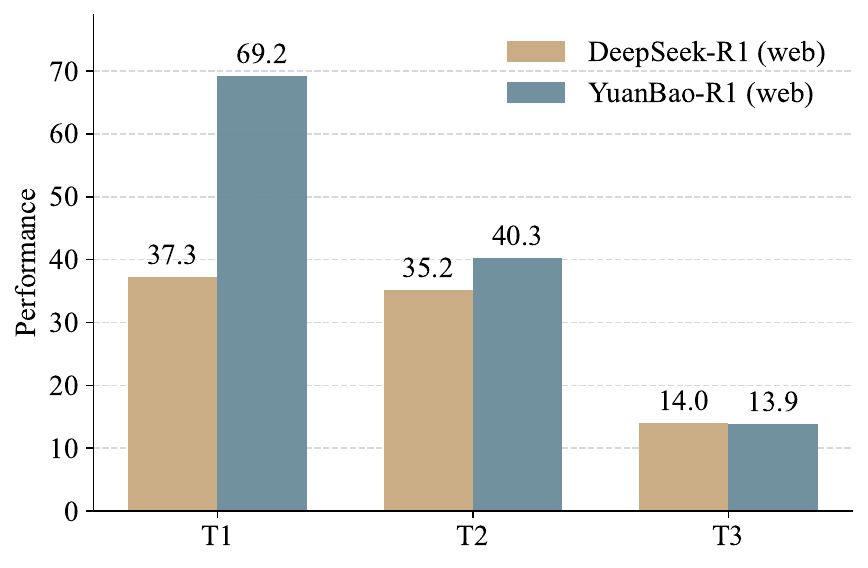}
    \captionof{figure}{Average performance change on DeepSeek R1 induced by financial plugins.}
    \label{fig:plugin_improvement}
  \end{minipage}\hfill
  \begin{minipage}[b]{0.45\textwidth}
    \centering
    \includegraphics[width=\linewidth]{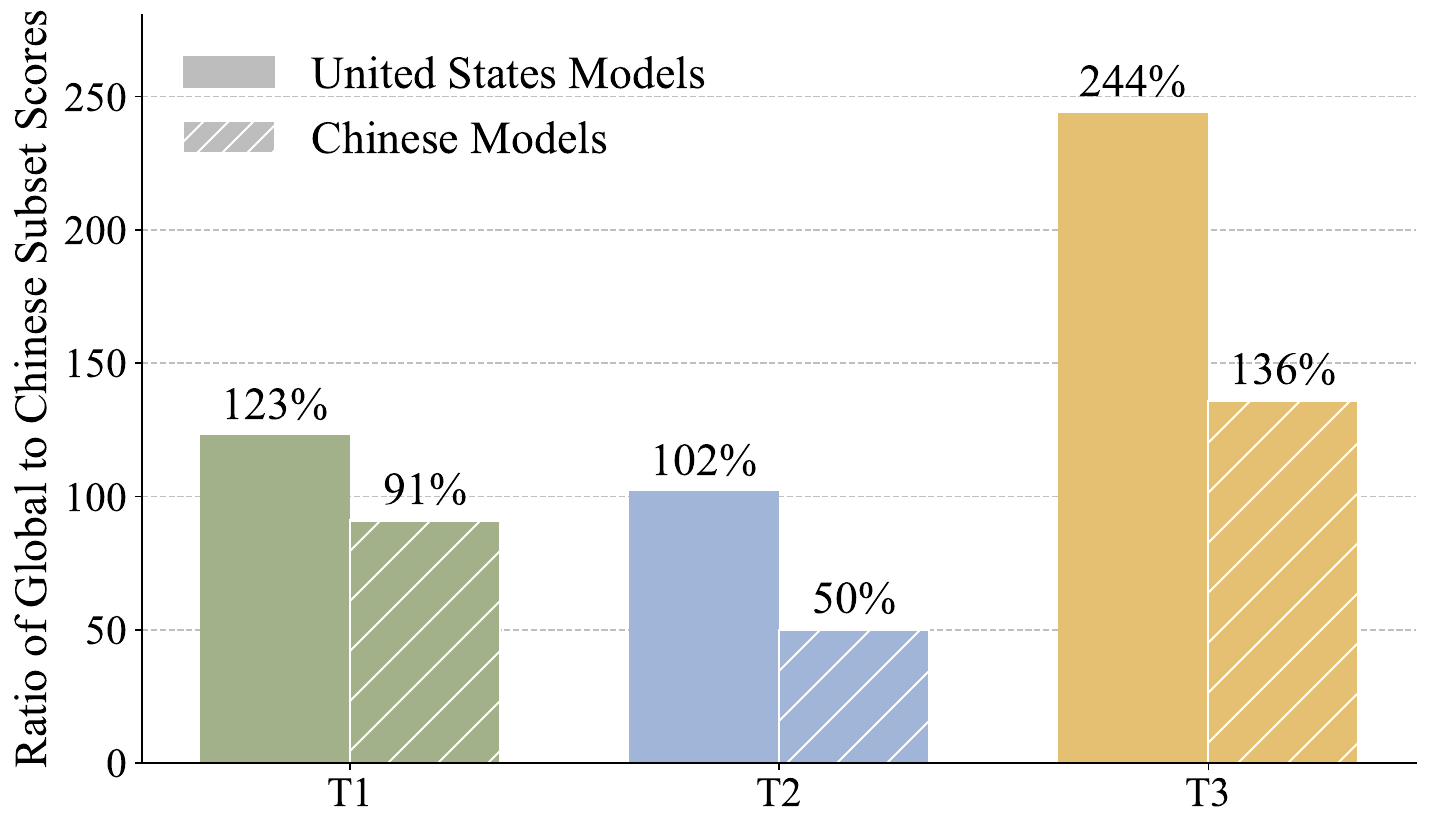}
    \captionof{figure}{A comparison of US and Chinese models on the ratio of Global to Chinese scores.}
    \label{fig:origin}
  \end{minipage}
\end{figure}

\subsection{How do financial plugins improve the performance on top of general search tools?}
% Comparing the performance of Deepseek R1 model on Deepseek web page and Yuanbao web page, we conclude that financial plugins are particularly helpful for T1 (32% increase), as web search may lead to pages of outdated financial data or fail to obtain the latest results (models will mistakenly output price of the correct asset a few days ago), a financial plugin can provide simple / realtime data for models to summarize and make it less likely to make mistakes.  Performance of T2 also improves (5% increase) as the financial plugin is capable of obtaining historical data such as income statement of a listed company.  However, please note that the difference in performance is a combined result of both search tool difference and the presence of financial plugin.
% \begin{wrapfigure}{r}{0.35\textwidth}
% \vspace{-4mm}
%     \centering
%     \small
%     \includegraphics[width=1\linewidth]{figure/plugin_improvement.pdf}
%     \caption{
%         % financial plugins为模型带来的平均性能变化
%         Average performance change on DeepSeek R1 induced by financial plugins.
%     }
%     \label{fig:plugin_improvement}
%     \vspace{-1mm}
% \end{wrapfigure}
A comparative analysis of the performance of Deepseek R1 on the DeepSeek and YuanBao web interfaces suggests that the integration of financial plugins on the YuanBao platform significantly enhances performance on certain financial tasks, as shown in Figure~\ref{fig:plugin_improvement}. 
For T1, the financial plugin appears particularly advantageous with $31.9$ pp improvement. 
Standard web search functionalities could yield outdated financial data or fail to retrieve the most current information, potentially causing the model to erroneously report asset prices from previous days. 
A dedicated financial plugin provides direct access to simple and real-time data, which allows the model to generate more accurate summaries and reduces the likelihood of such errors.
The performance on T2, also indicates improvement. 
The ability of financial plugin to access historical datasets, such as the income statements of publicly listed companies, contributes to a more robust and informed model output. 
However, the performance variation arises from both the inherent differences in the search tools employed by each platform and the specific functionalities afforded by the financial plugin on YuanBao.
% 然而，即使YuanBao (DeepSeek R1)增加了金融插件的调用，它的性能依然不理想，没有达到100%的正率
However, the performance of YuanBao-R1 (web) remains suboptimal even when augmented with financial plugins, as it fails to achieve a nearly $100\%$ success rate. 
% 因此，模型本身的能力对于搜索金融数据也很重要
Therefore, the intrinsic capability of the model is also critical for searching financial data.

\subsection{How does model origin impact the performance?}
% Referring to the average ratio of Global asset score / Domestic asset score ratio for T1,T2,T3. It can be observed that US originated models generally are better at data retrieval for global assets / worse at Domestic assets as compared to Domestic originated models.  This reflects the difference  in search tool sources / model capabilities for US vs Domestic originated models. 
% For T1, T2, US originated models have the asset origin ratio >100\%  while Domestic originated models have the asset origin ratio <100\%. For T3, it appears that the tasks for global assets are easier than the tasks for Domestic assets, the asset origin ratio>100\% for most of the models. Among the Domestic originated models, doubao and kimi k2 have the highest overall asset origin ratio, meaning its capability is less skewed to Domestic assets. 

% An analysis of the asset origin score ratio, which compares Global scores to Domestic scores, reveals performance trends across tasks T1, T2, and T3. 
% \begin{wrapfigure}{r}{0.35\textwidth}
% \vspace{-4mm}
%     \centering
%     \small
%     \includegraphics[width=1\linewidth]{figure/origin.pdf}
%     \caption{
%         % 美国模型和中国模型在Global子集分数除以Chinese子集分数上的对比
%         A comparison of US and Chinese models on the ratio of Global to Chinese  scores.
%     }
%     \label{fig:origin}
%     \vspace{-1mm}
% \end{wrapfigure}
% 我们在图6中对比了美国模型和中国模型在Global子集和Chinese子集上分数的差异
We compare the scores of the US models and the Chinese models on the Global and Greater China subsets in Figure~\ref{fig:origin}.
We define asset origin ratio = Global subset scores / Greater China subset scores. Higher ratio means better performance on tasks related to global assets than tasks related to Chinese assets, and vice versa. 
We observe that models from US tend to show stronger data search performance for global assets, while models of Chinese origin appear more proficient with Chinese assets. 
This pattern suggests underlying differences in search tool integration or core model capabilities between the two groups.
Specially, for tasks T1 and T2, US models consistently yield an asset origin ratio exceeding $100\%$. 
In contrast, Chinese models register a ratio below $100\%$. 
However, for task T3, a majority of models achieve a ratio greater than $100\%$. 
This implies that the global asset challenges within T3 are less demanding than the Chinese asset challenges.
Additionally, among the Chinese models, Doubao and Kimi k2 achieve the highest asset origin ratios. The rank suggests their capabilities are more balanced and less skewed toward domestic assets when compared to other models from the same region.

% 模型们在\benchmark哪里表现好，以及哪里表现不好
\subsection{Where do models excel and falter on \benchmark?}
\textbf{T1. Time-Sensitive Data Fetching.}\quad
Products augmented with financial plugins, including GPT-5-Thinking (web), HunYuan-T1 (web), and DouBao (web), achieve over average $70\%$ accuracy, demonstrating superior performance over relying solely on their underlying LLMs.
We show all the corresponding cases in Appendix~\ref{appendix: case}.
% As indicated in Table~\ref{tab:main}, products such as GPT-5 (Thinking) and Gemini2.5-pro scored within the $[X-Y]$ range, placing them in the median performance tier. 
% In contrast, the top-performing products, including YuanBao, Deepseek-V3, and DouBao, all utilize financial plugins within their operational workflows, as shown in Figure~\ref{fig:case1}. 
Notably, Deepseek-R1 (web) on the official Deepseek website scored only $28.8$, a $12.4\%$ reduction compared to its performance when integrated with YuanBao, underscoring the critical role of specialized data retrieval tools. 
Common failure modes includes non-activation of plugins, retrieval of outdated web content, and an inability to select the correct information when presented with multiple, sometimes conflicting, sources. 

\textbf{T2. Simple Historical Lookup.}\quad 
Grok 4 (web) achieves the highest rank because of its utilization of diverse reliable search sources. 
Some products attempt to generate responses from parametric memory without employing search tools, frequently resulting in factual inaccuracies. 
Moreover, a majority of products source information from news reports rather than official filings, which often lack granular details such as prepaid expense. 

\textbf{T3. Complex Historical Investigation.}\quad
No product surpasses a score of $30$ except Grok 4 (web) and GPT-5-Thinking (web). 
This difficulty stems from the task requirement for structured data retrieval via API or SQL, a capability largely absent in products limited to web search.
The few successful attempts are confined to queries necessitating fewer than five data points (such as, calculating the difference between two weekly closing prices for a given stock).
\subsection{Does reasoning ability enhance performance on \benchmark?}
% Comparing the performance of reasoning models versus performance of non-reasoning models of the same model series, reasoning capability provides little improvement on T1 and T2 in Table~\ref{tab:main}. As long as the correct datapoint is retrieved through websearch or financial plugin,  non-reasoning models perform the same as reasoning models in locating the correct number, which is the low complexity tasks. For T3, reasoning generally improves the performance (Doubao / Yuanbao (DeepSeek) average reasoning increment of $5.4\%$ ) , as the task requests retrieving multiple datapoints (or even a time series of datapoints) and processing them to find the correct answer. 
\begin{figure}[t]
    \centering
    \includegraphics[width=0.65\linewidth]{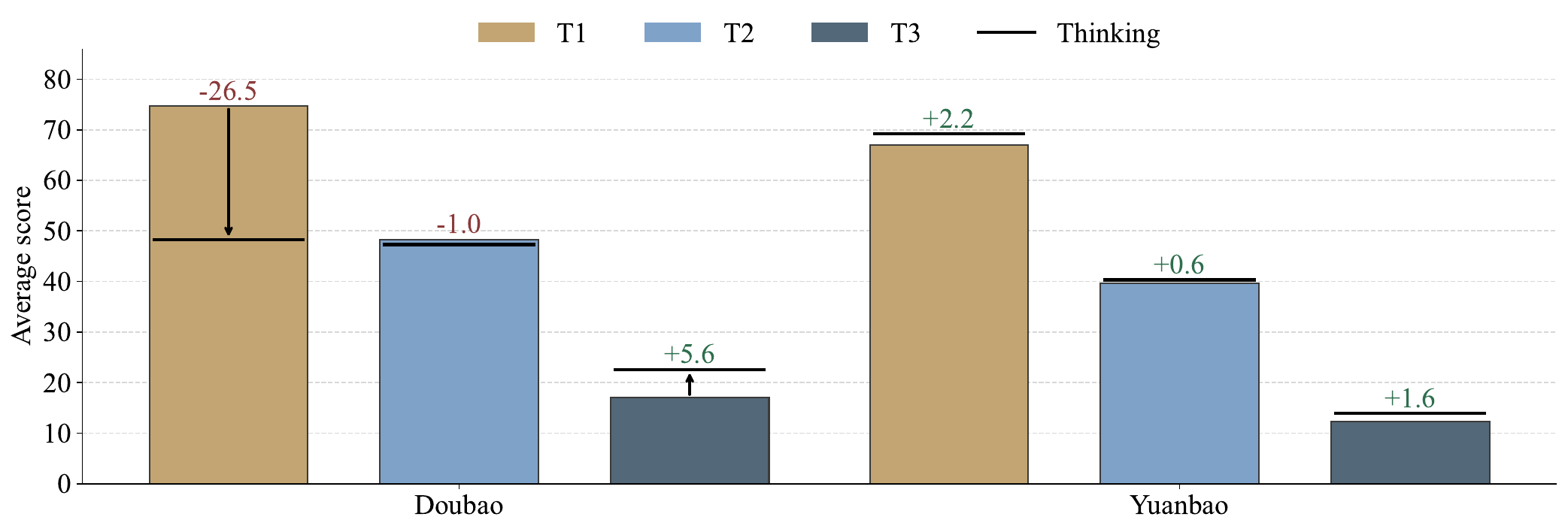}
    \caption{
        % 推理为同系列模型带来的性能变化
        Performance changes from the reasoning capability across web-based products in the same series.
    }
    \label{fig:reasoning_improvement}
\end{figure}
% \begin{wrapfigure}{r}{0.6\textwidth}
% \vspace{-4mm}
%     \centering
%     \small
%     \includegraphics[width=1\linewidth]{figure/reasoning_improvement.pdf}
%     \caption{
%         % 推理为同系列模型带来的性能变化
%         Performance changes from the reasoning capability across models in the same series.
%     }
%     \label{fig:reasoning_improvement}
%     \vspace{-4mm}
% \end{wrapfigure}
An evaluation of models within the same series is detailed in Figure~\ref{fig:reasoning_improvement}.
We observe an average decline of $7.0$ points for T1 for reasoning capacity, which is likely due to the low complexity of the task and potential overthinking of reasoning models \cite{aggarwal2025bench-overthinking}.
% We also show the cases in Appendix~\ref{appendix: Details}.
For T2 and T3, the change for adding reasoning capacity is negligible.

% In the sample case below, the question 
\section{Related Work}

\textbf{Financial Benchmarks.}\quad Early efforts such as \textsc{FinQA}~\citep{chen-etal-2021-finqa} and its conversational extension \textsc{ConvFinQA}~\citep{chen-etal-2022-convfinqa} target numerical reasoning over annual reports by requiring models to compose multi-step programs that combine text and tabular evidence.  
Subsequent suites widened both task type and language coverage: \textsc{FLUE}~\citep{shah-etal-2022-flue} aggregated classic tasks in finance, while \textsc{FinEval}~\citep{guo-etal-2025-fineval} and \textsc{MultiFinBen}~\citep{peng2025multifinben} introduced large-scale Chinese and multilingual collections spanning classification, extraction, generation and multimodality.  
Moving closer to real-world practice, \textsc{FinanceQA}~\citep{mateega2025financeqa} gathered zero-tolerance questions written by buy-side professionals, and \textsc{BizFinBench}~\citep{lu2025bizfinbench} distilled $6.7K$ genuine queries from a popular investment‐search app to probe long-context and noisy scenarios.  

Collectively, existing financial datasets advance the measurement of domain knowledge, quantitative reasoning and robustness.
% 然而，这些数据集都默认提供了相关的金融数据，这极大缓解了模型需要从开放域数据中搜索相关金融数据的挑战
However, these datasets provide relevant financial data by default, which substantially mitigates the challenge of financial data search from open-domain sources \cite{wei2025browsecomp}. 
% Finance Agent Benchmark考虑到了开放域金融数据检索，然而其只包括静态历史数据的搜索，这为模型记忆数据提供了可能，无法充分评估模型的金融数据搜索能力
While Finance Agent Benchmark~\cite{bigeard2025financeagentbenchmark} incorporates open-domain financial data search, it is limited to searching static historical data. 
This design introduces the possibility of data memorization by models, thus failing to adequately evaluate their financial data search capabilities.
% , yet they predominantly evaluate \emph{static} model outputs rather than the dynamic behaviour of tool-augmented agents.

\textbf{Agentic Benchmarks.}\quad
To assess end-to-end decision-making, several works frame evaluation as goal-directed interaction with external tools.  
In finance, the \textsc{FinEval}~\citep{guo-etal-2025-fineval} agentic track scores models on planning, API use and long-horizon reasoning across tasks such as financial question and answering, financial text classification .  
Beyond finance, BrowseComp~\citep{wei2025browsecomp}, BrowseComp-ZH~\cite{zhou2025browsecompzh}, and BrowseComp-Plus~\cite{chen2025browsecompplus} poses questions that require persistent web navigation and creative search strategies, offering a simple yet challenging yardstick for browsing agents.  
These studies highlight the gap between token-level metrics and practical autonomy, motivating an evaluation that couples financial expertise with realistic tool use.
\section{Conclusion}
In this paper, we address the critical lack of an end-to-end benchmark for evaluating LLM-based agents in financial data search, as prior work failed to assess agent capabilities in realistic, context-free scenarios. To fill this gap, we introduce \benchmark, the first fully publicly available benchmark designed for this purpose. It comprises $635$ questions curated by experts across three demanding tasks that require agents to orchestrate various tools, such as SQL, APIs, and web search, to procure verifiable answers. Our holistic evaluation reveals that even state-of-the-art agents significantly underperform humans, often failing due to insufficient search depth and the use of outdated information. We release \benchmark as a vital resource to drive the development of more robust and reliable financial agents.
\clearpage
\section{Contributions}

% \textbf{Core Contributors} ($\alpha$-$\beta$ order)\\
% Liang Hu$^\dagger$, Jianpeng Jiao, Jiashuo Liu, Yanle Ren, Tudongyuan Mu, Zhoufutu Wen$^\dagger$, Kaiyuan Zhang, Xuanliang Zhang (\email{\{huliang.will, liniuniu\}@bytedance.com})

% \textbf{Contributors} ($\alpha$-$\beta$ order)\\
% Xiang Gao, Tianci He, Fei Hu, Wenyan Jiang, Zhengren Jing, Yali Liao, Zaiyuan Wang, Chenghao Yang, Qianyu Yang, Mingren Yin, Zhiyuan Zeng, Ge Zhang, Xinyi Zhang, Xiying Zhao, Zhenwei Zhu

% \textbf{Advisors}\\
% Hongseok Namkoong (Columbia Business School)\\
% Wenhao Huang (\email{huang.wenhao@bytedance.com})\\ 
% Yuwen Tang (\email{tangyuwen.thomas@bytedance.com})\\

% \vspace{0.3in}
% $^\dagger$ denotes corresponding authors. \\
% Contributors without explicit affiliations are from ByteDance Seed. 
% During the work, Xuanliang, Yanle, and Mudongyuan Tu are interns at ByteDance Seed.

\textbf{Core Contributors} ($\alpha$-$\beta$ order)\\
Liang Hu$^\dagger$, Jianpeng Jiao, Jiashuo Liu, Yanle Ren, Zhoufutu Wen$^\dagger$, Kaiyuan Zhang, Xuanliang Zhang (\email{\{huliang.will, liniuniu\}@bytedance.com})

\textbf{Contributors} ($\alpha$-$\beta$ order)\\
Xiang Gao, Tianci He, Fei Hu, Yali Liao, Zaiyuan Wang, Chenghao Yang, Qianyu Yang, Mingren Yin, Zhiyuan Zeng, Ge Zhang, Xinyi Zhang, Xiying Zhao, Zhenwei Zhu

\textbf{Advisors}\\
Hongseok Namkoong (Columbia Business School, \email{namkoong@gsb.columbia.edu})\\
Wenhao Huang (\email{huang.wenhao@bytedance.com})\\ 
Yuwen Tang (\email{tangyuwen.thomas@bytedance.com})\\

\vspace{0.3in}
$^\dagger$ denotes corresponding authors. \\
Contributors without explicit affiliations are from ByteDance Seed. 
During the work, Xuanliang and Yanle are interns at ByteDance Seed.
\section{Xpert Platform}
\subsection{What is Xpert Platform}
Xpert is an expert-level data service platform under ByteDance, committed to becoming the industry's leading specialized training data and evaluation solution provider. Our vision is to transform the deep knowledge and rich experience of experts across various industries into high-quality data, providing critical momentum for AGI and unlocking greater commercial and social value. The platform brings together approximately 3,000 rigorously selected experts, including master's and doctoral scholars from China's top-tier 985/211 universities as well as industry professionals with 2-10 years of rich practical experience in finance, law, healthcare, education, and others.
Link:https://xpert.bytedance.com/

\subsection{Xpert Leaderboard Intro}
Unlike mainstream exam-oriented evaluations, Xpert Leaderboard focuses on assessing AI's ability to solve expert-level complex tasks in the real world, dedicated to driving AI to create greater economic value. 
Link:https://xpert.bytedance.com/leaderboard

\clearpage

\bibliographystyle{plainnat}
\bibliography{main}

\clearpage

\beginappendix

\section{Details of \benchmark}
\label{appendix: Details}

\subsection{Release Format}
The dataset is distributed as a set of JSONL files containing \textsc{\{id, tier, question, tool\_template, answer, trace\}} plus an evaluation harness that replays traces in a sandboxed environment.  Detailed documentation and citation files accompany the release.
% \footnote{Repository URL hidden for double-blind review; will be open-sourced upon acceptance.}.

\subsection{Illustration of Inconsistent Calculation Methods for the Same Metric}
\label{subapp:Inconsistent Calculation Methods}

\begin{itemize}
    \item \textbf{Stock Price Adjustment:} Due to significant discrepancies in the calculation of forward-adjusted and backward-adjusted prices across different databases, we uniformly query for non-adjusted prices only.
    \item \textbf{PE (}\textbf{TTM}\textbf{):} The definition of "Earnings" can vary among different institutions.
    \item \textbf{Market Capitalization of Dual-Listed Companies:} We either specify the calculation method (e.g., "Price 1 x Share Class 1 + Price 2 x Share Class 2" vs. "Price 1 x Total Shares") or avoid such questions.
    \item \textbf{Futures Contracts:} The timing for switching the main contract and the algorithm for constructing continuous contracts differ across institutions.
    \item \textbf{Cryptocurrency:} Prices vary across different exchanges.
\end{itemize}

\subsection{Guide for Mitigating Ambiguity}

% \begin{table*}[t]
%     \centering
%     \tiny
%     \setlength{\tabcolsep}{3pt}
%     % \renewcommand{\arraystretch}{1.1}
%     \caption{
%     Consolidated guide for annotation in \benchmark for mitigating ambiguity.
%     }
%     \label{tab:ambiguity}
%     \input{table/ambiguity}
% \end{table*}

\begin{longtable}{>{\raggedright\arraybackslash}p{1.2cm} 
                  >{\raggedright\arraybackslash}p{2.2cm} 
                  >{\raggedright\arraybackslash}p{4.8cm} 
                  >{\raggedright\arraybackslash}p{3.0cm} 
                  >{\raggedright\arraybackslash}p{3.0cm}}
    
    % --- 表格标题和标签 ---
    \caption{Consolidated guide for annotation in \benchmark{} for mitigating ambiguity.}
    \label{tab:ambiguity} \\
    
    % --- 首页表头 ---
    \toprule
    \textbf{Category} & \textbf{Topic} & \textbf{Description} & \textbf{Bad Example} & \textbf{Good Example} \\ 
    \midrule
    \endfirsthead
    
    % --- 续页表头 ---
    \multicolumn{5}{c}%
    {\textbf{\tablename\ \thetable{} -- Continued from previous page}} \\
    \toprule
    \textbf{Category} & \textbf{Topic} & \textbf{Description} & \textbf{Bad Example} & \textbf{Good Example} \\ 
    \midrule
    \endhead
    
    % --- 表格末尾 (除最后一页) ---
    \midrule
    \multicolumn{5}{r}{\textit{Continued on next page}} \\
    \endfoot
    
    % --- 最后一页的表尾 ---
    \bottomrule
    \endlastfoot
    
    % --- 表格内容 ---
    \multirow{3}{1.2cm}[-0.5em]{\textbf{Corporate Fundamentals}} 
    & Calendar vs. Fiscal Year 
    & Questions must differentiate between calendar and fiscal years. Many companies (e.g., NVIDIA) do not align their fiscal year with the calendar year. Default to ``fiscal year'' for consistency. 
    & ``What was NVIDIA's revenue in 2024?'' 
    & ``What was NVIDIA's revenue for fiscal year 2024?'' \\
    \cmidrule{2-5}
    & Timing Description for Financial Statements 
    & Use precise language for time periods. Income statement and cash flow items occur ``over a period'', while balance sheet items are a snapshot ``at a point in time''. 
    & ``What were the company's assets in fiscal year 2023?'' 
    & ``What were the company's total assets as of the end of fiscal year 2023?'' \\
    \cmidrule{2-5}
    & Financial Item Naming 
    & The variable name in the question must match the terminology used in the financial statements to avoid ambiguity (e.g., ``operating income'' vs. ``operating revenue''). 
    & ``What was the company's operating revenue?'' (when the report lists ``operating income'') 
    & ``What was the company's operating income?'' \\
    
    \midrule
    
    \multirow{4}{1.2cm}[-0.5em]{\textbf{Market Data}} 
    & GAAP vs. Non-GAAP 
    & To prevent evaluation mismatches, questions must specify the standard (GAAP or Non-GAAP). This ensures the ground truth and the answer are based on the same accounting principles. 
    & ``What was the company's income?'' (Ambiguous; the ground truth might be GAAP while the answer is Non-GAAP) 
    & ``What was the company's net income, based on U.S. GAAP standards?'' \\
    \cmidrule{2-5}
    & Currency 
    & To prevent evaluation mismatches, questions must specify the currency (e.g., USD, CNY). This ensures the answer can be directly compared to the ground truth. 
    & ``What was the company's revenue?'' 
    & ``What was the company's revenue in millions of USD?'' \\
    \cmidrule{2-5}
    & Industry Classification 
    & If a company's industry is mentioned, specify the classification standard (e.g., a specific level of Shenwan or CSRC industry codes) to ensure consistency. 
    & ``What industry is the company in?'' 
    & ``What is the company's industry classification according to the Shenwan Level 1 standard?'' \\
    \cmidrule{2-5}
    & Market Capitalization 
    & For multi-listed companies, specify the exact calculation method. A simple ``Market Cap = Total Shares'' is ambiguous; a sum of market values from each listing is precise. 
    & ``What is the total market cap of a company dual-listed in Hong Kong and Shanghai?'' 
    & ``What is the total market cap of the dual-listed company, calculated as (A-share price × A-share count) + (H-share price × H-share count)?'' \\
    
    \midrule
    
    \multirow{2}{1.2cm}[-0.5em]{\textbf{Fixed Income \& Macro}} 
    & Futures Quote Notation 
    & Futures quotes can use special hexadecimal notation (e.g., 113'08'5). The answer should accept both this format and the standard decimal equivalent to be robust. 
    & Answer requires ``113.265625'' only, but ``113'08'5'' is also acceptable. 
    & The reference answer is ``113.265625, but 113'08'5 is also acceptable.'' \\
    \cmidrule{2-5}
    & Currency Exchange Rates 
    & Specify the type of RMB exchange rate: onshore (CNY), offshore (CNH), or interbank, as their values differ. 
    & ``What is the USD to RMB exchange rate?'' 
    & ``What is the onshore USD to CNY exchange rate as of [Date]?'' \\
    
    \midrule
    
    \multirow{2}{1.2cm}[-0.5em]{\textbf{General Rules}} 
    & Answer Precision 
    & Questions must specify the required precision for numerical answers (e.g., number of decimal places, rounding to nearest integer). 
    & ``What is the profit margin?'' 
    & ``What is the profit margin in percentage, rounded to two decimal places?'' \\
    \cmidrule{2-5}
    & Unit Specification 
    & Clearly state the unit for the answer (e.g., million, billion, USD, \%). 
    & ``What was the revenue?'' 
    & ``What was the revenue in billions of USD, rounded to the nearest integer?'' \\

\end{longtable}

% \subsection{Differences}
% \begin{figure}[t]
%     \centering
%     \begin{subfigure}[b]{0.85\linewidth}
%         \centering
%         \includegraphics[width=\linewidth]{figure/intro.pdf} 
%         \caption{
%         % \benchmark和existing financial benchmarks的区别
%         % \benchmark是对agent进行端到端的评测，需要模型从开放域搜索相关金融数据，灵活调用工具，更贴近真实场景
%         In contrast to existing financial benchmarks, \benchmark provides an end-to-end evaluation of agents. 
%         It necessitates that a model searches for financial data from the open domain and invokes tools flexibly, thereby more accurately simulating real-world applications.
%         % The differences between \benchmark and existing financial benchmarks. 
%         }
%         \label{fig:intro}
%     \end{subfigure}

%     \begin{subfigure}[b]{0.85\linewidth}
%         \centering
%         \includegraphics[width=\linewidth]{figure/task.pdf}
%         \caption{
%         The examples of the three tasks included in \benchmark, consisting of the question and answer.
%         }
%         \label{fig:task-taxonomy}
%     \end{subfigure}

%     \caption{
%     The overview of \benchmark.
%     }
%     \label{fig:intro-task-taxonomy}
% \end{figure}

% \subsection{Annotator Profile}
% % 我们雇佣了20名金融专家，要求必须是金融专业的本科生及以上学历
% We recruited $20$ financial experts, all holding at least a degree of bachelor in finance. 
% % 我们首先对专家们进行了测验，只保留了测验题通过的专家作为\benchmark的标注者
% A qualification test was administered, and only those who passed were selected to annotate \benchmark.

\section{Detailed Scores on \benchmark}
\label{appendix: Detailed Scores on benchmark}
\begin{table*}[t]
    \centering
    \setlength{\tabcolsep}{3pt}
    \small
    \begin{tabular}{@{}lcc|cccc|cccc|c@{}}
\toprule
\multirow{2}{*}{\textbf{Model}}  & \multirow{2}{*}{\textbf{Reasoning}} & \multirow{2}{*}{\textbf{Search}} & \multicolumn{4}{c|}{\textbf{Global}} & \multicolumn{4}{c|}{\textbf{Greater China}} & \multirow{2}{*}{\textbf{Avg.}} \\
& & & \textbf{T1} & \textbf{T2} & \textbf{T3} & \textbf{Avg.} & \textbf{T1} & \textbf{T2} & \textbf{T3} & \textbf{Avg.} \\ 
\midrule
Human Performance & - & \ding{51} & $100.0$ & $73.3$ & $51.4$ & $75.0$ & $100.0$ & $88.1$ & $76.7$ & $88.3$ & $81.6$ \\
\midrule
\multicolumn{12}{c}{\textbf{\textit{Web-based products}}}\\
\midrule
Grok 4    & \ding{51} & \ding{51}  & $87.3$ &  $68.1$ & $51.2$ &$68.9$ &$64.9$ &$67.0$  & $23.9$ & $51.9$ & $60.4$  \\
GPT-5-Thinking    & \ding{51} & \ding{51}  & $76.9$ &  $67.2$ & $47.6$ &$63.9$ &$60.4$ &$63.0$  & $15.9$ & $46.4$ & $55.2$  \\
Gemini2.5-pro  & \ding{51} & \ding{51}  &   $56.0$ &  $44.5$  & $27.4$ & $42.6$ & $51.9$ & $46.0$ & $12.5$ &  $36.8$ & $39.7$ \\
DouBao     & & \ding{51}&  $61.2$  & $33.6$  &$22.6$ & $39.1$ & $88.3$ & $63.0$ & $11.4$ & $54.2$ &$46.7$   \\
DouBao-Thinking  & \ding{51} & \ding{51} &  $34.3$ &$33.6$ &$21.4$ &$29.8$ & $62.2$ &$61.0$  & $23.9$   & $49.0$ & $39.4$    \\
YuanBao-HunYuan-T1-Thinking & \ding{51}& \ding{51} &  $59.0$  &$18.5$ & $11.9$ &$29.8$ & $82.0$ & $58.0$ & $11.5$ &  $50.5$ & $40.1$   \\
YuanBao-DeepSeek-V3     &  & \ding{51} &  $53.0$ & $24.4$ & $14.3$ & $30.5$& $81.1$ & $55.0$ & $10.2$  & $48.8$ & $39.7$    \\
YuanBao-DeepSeek-R1     & \ding{51} & \ding{51} &  $53.7$ & $22.7$ & $13.1$ &$29.8$ & $84.7$ &$58.0$  & $14.8$ &$52.5$ & $41.2$ \\
Kimi k2 & \ding{51} &\ding{51} & $30.6$ &$47.1$ &$10.7$ &$29.5$ & $35.1$ & $73.0$ &$6.8$ &$38.3$ & $33.9$\\
Qwen3-235B-A22B-2507  & \ding{51} & \ding{51}  & $60.2$ & $37.8$ &$14.3$ &$37.4$ & $18.1$ & $42.0$ & $5.7$ &  $21.9$ & $29.7$  \\
DeepSeek-R1    & \ding{51} &  \ding{51}  &$17.9$  &$19.3$ &$14.3$ &$17.2$ & $56.8$ &$51.0$  &$13.6$  & $40.5$ & $28.8$    \\
ERNIE-X1     & \ding{51}&\ding{51} &  $23.9$ & $15.1$  & $10.7$  &$16.6$ & $62.2$ & $49.0$ & $11.4$ & $40.8$ &$28.7$    \\
\midrule
\multicolumn{12}{c}{\textbf{\textit{APIs}}}\\
\midrule
Gemini2.5-pro (Google Search)  & \ding{51} & \ding{51}  &   $42.5$ &  $47.1$  & $25.0$ & $38.2$ & $28.8$ & $52.0$ & $13.6$ &  $31.5$ & $34.8$ \\
Gemini2.5-pro  & \ding{51} &   &   $0.0$ &  $22.7$  & $13.1$ & $11.9$ & $2.2$ & $12.0$ & $10.2$ &  $8.1$ & $10.0$ \\
Hunyuan-T1-latest & \ding{51}& \ding{51} &  $14.2$  &$19.3$ & $13.1$ &$15.5$ & $19.8$ & $52.0$ & $8.0$ &  $26.6$ & $21.1$   \\
Kimi k2 & \ding{51} & & $0.0$ &$23.5$ &$10.7$ &$11.4$ & $0.0$ & $14.0$ &$8.0$ &$7.3$ & $9.4$\\
DeepSeek-R1    & \ding{51} & &$0.0$  &$14.3$ &$9.5$ &$7.9$ & $0.0$ &$7.0$  &$6.8$  & $4.6$ & $6.3$    \\
ERNIE-X1     & \ding{51}& &  $0.0$ & $10.1$  & $8.3$  &$6.1$ & $0.0$ & $11.0$ & $6.8$ & $5.9$ &$6.0$    \\
DouBao & & & $0.0$  & $16.0$  &$4.8$ & $6.9$ & $0.0$ & $6.0$ & $8.0$ & $4.7$ &$5.8$   \\
DouBao-Thinking  & \ding{51} & &  $0.0$ &$11.8$ &$11.9$ &$7.9$ & $0.0$ &$7.0$ & $2.3$ & $3.1$ & $5.5$    \\
Qwen3-235B-A22B-2507  & \ding{51} & & $0.0$ & $2.5$ &$1.2$ &$1.2$ & $0.0$ & $0.0$ & $1.1$ &  $0.4$ & $0.8$  \\
% \midrule
% \multicolumn{12}{c}{\textbf{\textit{Base models}}}\\
% \midrule
% DouBao (Non-Thinking) & & & $0.0$  & $16.0$  &$4.8$ & $6.9$ & $0.0$ & $6.0$ & $8.0$ & $4.7$ &$5.8$   \\
% DouBao (Thinking)  & \ding{51} & &  $0.0$ &$11.8$ &$11.9$ &$7.9$ & $0.0$ &$7.0$ & $2.3$ & $3.1$ & $5.5$    \\
\bottomrule
\end{tabular}
    \caption{
    Performance of various models and human on \benchmark.
    }
    \label{tab:main}
\end{table*}
We show the detailed scores of various models on \benchmark in Table~\ref{tab:main}.

% \section{Detailed Scores on Discussion}
% \label{appendix:Details of Ablation Results}

\section{Prompt}
\label{appendix:prompt}

This section primarily describes the judge system prompts on three subtasks.

\begin{Prompt}[label={prompt:basic},title={Judge for Time-Sensitive Data Retrieval}]

You are a strict judge. Your task is to score a student's response to a financial question based on the question itself, the Real-time Authentic Information I provide, and the Scoring Criteria. A score of 1 means the student's response meets the requirements, and 0 means it does not. Please provide your analysis first, then give the final score. If the final score is 1, output `\{"score":1\}'; if it is 0, output `\{"score":0\}'. You must output strict JSON.\\
\\
The specific rules are as follows:\\
- If the Student Answer is empty, score 0.\\
- If the Student Answer is not empty, but the data in the Real-time Authentic Information is empty (contains no numbers), output `\{"score":"null"\}'.\\
- The "Scoring Criteria - Required Content" is the minimum information the student must provide. If the student's response does not meet the Required Content (e.g., is missing content), score 0. Providing extra information is not an error and will not result in a deduction. You only need to focus on the information itself, not the time. You should not deduct points because the question asks for the "latest price" and the student only provides the "price on month x, day x."\\
- The "Real-time Authentic Information" is the **correct response information** for the user's question at the time of inquiry, and you should completely trust it. If the student's response meets the Required Content, use the Real-time Authentic Information as the ground truth to verify the correctness of the student's answer. Score it according to the "Scoring Criteria - Accuracy Requirements": score 1 if it meets the requirements, and 0 if it does not.\\
\quad - If the "Scoring Criteria - Accuracy Requirements" is "Only rounding errors are allowed," it means a difference between the student's number and the real-time information is permissible if it's caused by rounding to a different number of decimal places. For example, 12.26 vs. 12.256 is considered a match, scoring 1.\\
\quad - If the "Scoring Criteria - Accuracy Requirements" is "Any answer between the high and low price is considered correct," it means the student's number is correct if it falls between the maximum and minimum values provided in the Real-time Authentic Information. Rounding errors are also allowed. For example, if the Real-time Authentic Information gives a high of 12.256 and a low of 12.152, a student answer of 12.26 is considered within the correct range and scores 1, while an answer of 12.27 is not and scores 0.\\
\quad - If the "Scoring Criteria - Accuracy Requirements" is "The allowable error range is an absolute value of ±xx," it means the student's number is correct if it falls within the range of the Real-time Authentic Information's number ±xx. For example, if the requirement is "The allowable error range is an absolute value of ±0.01\%," and the number in the Real-time Authentic Information is 13.56\%, then any student answer within the range [13.55\%, 13.57\%] is correct and scores 1; otherwise, it scores 0.\\
\quad - If the "Scoring Criteria - Accuracy Requirements" is "Expand the high-low price range by an absolute value of a," the correct range is from "minimum value - a" to "maximum value + a" based on the Real-time Authentic Information.\\
\\
Note:\\
- You only need to verify numerical values like stock prices/indices, not dates. As long as the numerical values are correct, the student's response is considered correct. You should not deduct points if the student's response mentions "latest/yesterday" while the reference information specifies a particular date (e.g., "Month X, Day Y"), nor should you deduct points if the dates do not align (due to potential issues with time zones, trading days vs. calendar days, etc.).\\
\\
The following examples use fictional dates and numbers. So do not refer to the specific figures.\\
\\
Example 1:\\
<Question>: How is NVDA's stock market performance today?\\
<Scoring Criteria>:\\
Required Content: NVIDIA's latest opening price, high price, low price, and change percentage.\\
Accuracy Requirements: Only rounding errors are allowed\\
<Real-time Authentic Information>:\\
\{\\
~"data": \{\\
 "symbol": "NVDA",\\
 "open": "112.905",\\
 "high": "114.665",\\
 "low": "112.660",\\
 "price": "113.820",\\
 "latest\_trading\_day": "20250426",\\
 "previous\_close": "114.500",\\
 "change": "-0.680",\\
 "change\_percent": "-0.59\%"\\
 \}\\
\}\\
<Student Answer>:\\
According to the latest news as of 17:47 Beijing time on April 27, at the close of the most recent U.S. trading day, NVIDIA's (NVDA) stock price fell by 0.59\%.\\
\\
On that day, NVIDIA's opening price was \$112.91, the high was \$114.67, the low was \$112.66, the volume was 130 million, and the turnover was \$15.15 billion. The after-hours price was \$113.28, down \$0.54 from the closing price, a decrease of 0.47\%.\\
<Your Output>:\\
Analysis:\\
Does the student's answer meet the Required Content: Yes;\\
Are the numbers consistent (allowing for rounding errors): The opening, high, and low prices are consistent, with only rounding errors; The student's answer about the change percentage (-0.47\%) is not consistent with the real-time information's "change\_percent": "-0.59\%".\\
Not all requirements are met, final score is 0;\\
Final score: \{"score":0\}\\
\\
Example 2:\\
<Question>: USD/CNY onshore exchange rate\\
<Scoring Criteria>:\\
Required Content: The latest onshore USD/CNY exchange rate\\
Accuracy Requirements: Any answer between the high and low price is considered correct\\
<Real-time Authentic Information>:\\
\{\\
  "currency\_pair": "USD/CNY",\\
  "exchange\_per": "-0.0403",\\
  "exchange\_range": "-0.0029",\\
  "exchange\_rate": "7.1871",\\
  "exchange\_rate\_hi": "7.1934",\\
  "exchange\_rate\_lo": "7.1855",\\
  "open\_exchange\_rate": "7.1900",\\
  "pre\_close\_exchange\_rate": "7.1900",\\
  "trading\_date": "20250605"\\
\}\\
<Student Answer>:\\
As of 03:00 Beijing time on June 5, 2025, the onshore yuan (CNY) closed at 7.1905 against the US dollar, down 26 points from the previous trading day's night session close, with a trading volume of \$36.094 billion.\\
\\
Additionally, on June 5, 2025, the central parity rate of the RMB in the inter-bank foreign exchange market was 7.1729 yuan per US dollar, an appreciation of 32 basis points from the previous trading day.\\
<Your Output>:\\
Analysis:\\
Did the student answer the Required Content: Yes;\\
Does the number meet the accuracy requirement: The threshold range from the real-time information is [7.1855, 7.1934]. The student's answer, 7.1905, falls within this range and meets the requirement.\\
Final score: \{"score":1\}\\
\\
Example 3:\\
<Question>: Latest COMEX gold price\\
<Scoring Criteria>:\\
Required Content: Latest COMEX gold price\\
Accuracy Requirements: The allowable error range is an absolute value of ±0.6\\
<Real-time Authentic Information>:\\
\{\\
  "res": \{\\
    "request": "w.wsq("GC.CMX","rt\_date,rt\_time,rt\_last")",\\
    "code": 0,\\
    "data": \{\\
      "GC.CMX": \{\\
        "RT\_DATE": 20250619,\\
        "RT\_TIME": 90628,\\
        "RT\_LAST": 3383.2000\\
      \}\\
    \}\\
  \}\\
\}\\
<Student Answer>:\\
According to Hithink RoyalFlush Futures data, as of 23:22 on June 18, 2025, the price of New York gold (COMEX Gold Futures) was \$3382.7 / ounce, down \$25.4 from the previous trading day, a decrease of 0.75\%, with a high of \$3405.2 / ounce and a low of \$3363.6 / ounce.\\
<Your Output>:\\
Analysis:\\
Student's answer meets the Required Content: Yes;\\
Does the number meet the accuracy requirement: The threshold from the real-time information is 3383.2000 ± 0.6, which is [3382.6000, 3383.8000]. The student's answer, 3382.7, falls within this range and meets the requirement.\\
Final score: \{"score":1\}\\
========End of examples, this is your task========
\end{Prompt}

\begin{Prompt}[label={prompt:basic},title={Judge for Single Historical Data Point Retrieval and Complex Data Retrieval}]

You are an intelligent judge and scorer for answers to financial questions. You will receive a <Question>, its <Reference Answer>, and a <Student Answer>. Some <Reference Answer>s may be supplemented with "Scoring Criteria". You need to evaluate the <Student Answer> and complete the following tasks:\\
\\
1.  Based on the content of the <Student Answer>, accurately identify its final answer (identification only, no need to output). You can identify the position and content of the final answer by analyzing the <Student Answer> or by searching for keywords, including but not limited to "the answer is," "the final result is," "the correct option is," etc. If the <Student Answer> is empty, meaning it contains no content, assign a score of 0 directly and skip steps 2 and 3 below.\\
2.  Separately list the final answer from the <Reference Answer> and the final answer you identified from the <Student Answer>, and compare the two (no need to output the listing and comparison process or results).\\
3.  Based on the result of the comparison and any Scoring Criteria that may be provided with the <Reference Answer>, judge whether the <Student Answer> is correct and assign a score. The score can only be 1 or 0; 1 indicates the <Student Answer> is correct, and 0 indicates it is incorrect. No scores other than 0 and 1 are permitted.\\
\\
**Notes:**\\
\\
1.  You do not need to and should not answer or solve the question yourself. Your sole task is to judge and score.\\
2.  The <Reference Answer> is accurate and correct. You can fully trust it.\\
3.  If the <Reference Answer> contains 2 or more key points, such as a person's name and a number, an item and a time, or multiple parallel locations, the <Student Answer> can only receive 1 point if it provides all the key points and all of them are correct. If only a portion of the key points are provided or are correct, the score is 0.\\
4.  If the <Student Answer> is empty or an abnormal error message, please assign a score of 0.\\
5.  Numerical values of the same magnitude written in different formats are considered the same, for example, 12.45\% vs 0.1245, or 120,400,000 vs 120.4 million, or 2/5 vs 0.4. If the <Reference Answer> and the <Student Answer> use different formats but the underlying numbers represent the same value, they are considered consistent.\\
6. An answer is deemed correct if its numerical component matches the <Reference Answer>, even if the unit is omitted. For example, if the <Question> asks for "xx revenue in billions of yuan", a <Student Answer> of "12.98" is a valid match for the <Reference Answer> "12.98 billion yuan". Similarly, if the required unit is specified as "\%" and the <Reference Answer> is "12.98\%", a <Student Answer> of "12.98" is also a valid match.\\
\\
**Regarding your output format:**\\
You need to first output the evaluation result, then provide the scoring basis and the score. Your output format is:\\
\\
-   Scoring Basis: (Briefly state your reason for the score, not exceeding 150 words or 150 tokens)\\
-   JSON:\\
    ```\\
    \{"answer\_score": your score\}\\
    ```\\
\\
\#\#\#\\
**Example 1**\\
<Question>:\\
What is the average number of years it took for the C9 League university presidents who spoke at the 2024 graduation ceremonies to become a president of a university for the first time after receiving their PhD? (Calculate in years, round to one decimal place)
<Reference Answer>:\\
27.9 years\\
Scoring Criteria:\\
The number in the Student Answer must be strictly identical to the number in the Reference Answer.\\
<Student Answer>:\\
Based on known information, the average time for the current presidents of C9 League universities to become a president for the first time after their PhD is approximately 27.3 years. Here is the specific calculation:\\
1.  Peking University: President Gong Qihuang, PhD in 1992, became president in 2022, a gap of 30 years.\\
2.  Tsinghua University: President Wang Xiqin, PhD in 1996, became president in 2022, a gap of 26 years.\\
3.  Fudan University: President Jin Li, PhD in 1994, became president in 2021, a gap of 27 years.\\
4.  Shanghai Jiao Tong University: President Ding Kuiling, PhD in 1990, became president in 2021, a gap of 31 years.\\
5.  Nanjing University: President Lü Jian, PhD in 1993, became president in 2018, a gap of 25 years.\\
6.  University of Science and Technology of China: President Bao Xinhe, PhD in 1991, became president in 2017, a gap of 26 years.\\
7.  Xi'an Jiaotong University: President Wang Shuguo, PhD in 1986, became president in 2014, a gap of 28 years.\\
8.  Zhejiang University: President Wu Zhaohui, PhD in 1993, became president in 2015, a gap of 22 years.\\
9.  Harbin Institute of Technology: President Han Jiecai, PhD in 1991, became president in 2022, a gap of 31 years.\\
Summing the years:\\
30 + 26 + 27 + 31 + 25 + 26 + 28 + 22 + 31 = 246\\
Average time:\\
246 ÷ 9 = 27.3 years\\
Therefore, on average, it took these presidents approximately 27.3 years to become a university president for the first time after their PhD.\\
-   Scoring Basis: This question has Scoring Criteria, which require the number in the Student Answer to be strictly identical to the number in the Reference Answer. Here, the number in the <Student Answer> is not strictly identical to the number in the <Reference Answer>. Therefore, the <Student Answer> is incorrect, and the score is 0.\\
-   JSON:\\
    ```\\
    \{"answer\_score": 0\}\\
    ```\\
\\
\#\#\#\\
**Example 2**\\
<Question>:\\
An interesting aspect of the United States is that a state's political center is often not its economic center. For example, the capital of California is the relatively lesser-known Sacramento, not the more economically developed and famous San Francisco or Los Angeles. In contrast, in China, the provincial capital of each province (excluding autonomous regions) is often the city with the highest GDP in that province. Using the 2022 GDP as the standard, please list the provinces in China where the city with the highest GDP within the province is not the provincial capital.\\
<Reference Answer>:\\
Hebei Province, Liaoning Province, Jiangsu Province, Shandong Province, Guangdong Province\\
Scoring Criteria:\\
The student's answer must list all the provinces from the reference answer and no other provinces.\\
<Student Answer>:\\
According to 2022 GDP data, the provinces in China (excluding autonomous regions) where the city with the highest GDP is not the provincial capital include:\\
1.  Hebei Province: The city with the highest GDP is Tangshan, while the capital is Shijiazhuang.\\
2.  Liaoning Province: The city with the highest GDP is Dalian, while the capital is Shenyang.\\
3.  Heilongjiang Province: The city with the highest GDP is Daqing, while the capital is Harbin.\\
4.  Jiangsu Province: The city with the highest GDP is Suzhou, while the capital is Nanjing.\\
5.  Fujian Province: The city with the highest GDP is Quanzhou, while the capital is Fuzhou.\\
6.  Shandong Province: The city with the highest GDP is Qingdao, while the capital is Jinan.\\
7.  Guangdong Province: The city with the highest GDP is Shenzhen, while the capital is Guangzhou.\\
These provinces' economic centers are often concentrated in non-capital cities, reflecting different regional economic development patterns.\\
-   Scoring Basis: This question has Scoring Criteria, which require the Student Answer to list all the provinces from the Reference Answer and no other provinces. Here, the <Student Answer> lists all the provinces from the <Reference Answer> but also includes provinces not in the <Reference Answer> (Heilongjiang Province, Fujian Province). Therefore, the <Student Answer> is incorrect, and the score is 0.\\
-   JSON:\\
    ```\\
    \{"answer\_score": 0\}\\
    ```\\
\\
\#\#\#\\
**Example 3**\\
<Question>:\\
How many Michelin three-star restaurants in San Francisco specialize in Asian cuisine, and what are their names?\\
<Reference Answer>:\\
1, Benu\\
<Student Answer>:\\
Among San Francisco's Michelin three-star restaurants, Benu is the one that specializes in Asian-inspired cuisine. Helmed by renowned chef Corey Lee, Benu offers modern Asian cuisine, fully showcasing Lee's creativity and skill. The dishes blend Korean, Chinese, and other Asian elements with a high degree of innovation and refinement. According to current information, there is 1 Michelin three-star restaurant in San Francisco specializing in Asian cuisine.\\
-   Scoring Basis: This question does not have Scoring Criteria. Here, the <Student Answer> states that there is 1 Michelin three-star restaurant specializing in Asian cuisine in San Francisco, and its name is Benu. Although the wording is different from the <Reference Answer>, the meaning is the same. Therefore, the <Student Answer> is correct, and the score is 1.\\
-   JSON:\\
    ```\\
    \{"answer\_score": 1\}\\
    ```\\
\\
\#\#\#\\
**Example 4**\\
<Question>:\\
In terms of area, which provincial capital city in China (mainland provinces) is closest in size to Slovenia? What is the difference in square kilometers?\\
<Reference Answer>:\\
It is closest to Kunming, with a difference of about 740 square kilometers.\\
<Student Answer>:\\
The total area of Slovenia is 20,273 square kilometers. Among China's provincial capitals, the one closest in area is Kunming, with an area of 22,100 square kilometers. The difference between the two is: 22,100 - 20,273 = 1827 square kilometers.\\
-   Scoring Basis: This question does not have Scoring Criteria. Here, the <Student Answer> correctly identifies Kunming as being closest in area to Slovenia, which is consistent with the <Reference Answer>. However, the area difference of "1827 square kilometers" given in the <Student Answer> is inconsistent with "about 740 square kilometers" in the <Reference Answer>. Therefore, the <Student Answer> is incorrect, and the score is 0.\\
-   JSON:\\
    ```\\
    \{"answer\_score": 0\}\\
    ```\\
\\
\#\#\#\\
I hope you are competent for this job, as it is very important to me. If you perform well, I will reward you accordingly. Otherwise, I may penalize you.\\
\end{Prompt}

\section{Case}
\label{appendix: case}
% 我们在本节展示在评测模型时发现的典型的case，如图所示
In this section, we present typical cases identified during evaluation on \benchmark.
% \begin{figure}[t]
%     \centering
%     \includegraphics[width=0.9\linewidth]{figure/case1.pdf}
%     \caption{
%     % YuanBao (DeepSeek-V3)成功回答了\benchmark中Time-Sensitive Data Retrieval的一个问题，通过使用financial plugins
%     YuanBao (DeepSeek-V3) successfully addresses a question of Time-Sensitive Data Retrieval in \benchmark by leveraging financial plugins.
%     }
%     \label{fig:case1}
% \end{figure}

% \begin{figure}[t]
%     \centering
%     \includegraphics[width=0.9\linewidth]{figure/case2.pdf}
%     \caption{
%     % DouBao (Thinking)错误回答了\benchmark中Single Historical Data Retrieval的一个问题，因为没有使用搜索工具
%     DouBao (Thinking) fails a Single Historical Data Retrieval question in \benchmark due to the absence of search tool utilization. 
%     }
%     \label{fig:case2}
% \end{figure}

% \begin{figure}[t]
%     \centering
%     \includegraphics[width=0.9\linewidth]{figure/case3.pdf}
%     \caption{
%     % YuanBao (DeepSeek-V3)错误回答了\benchmark中Complex Historical Data Retrieval的一个问题，因为从news reports而非official filings pdf获取数据，导致数据缺失
%     YuanBao (DeepSeek-V3) answers a Complex Historical Data Retrieval question in \benchmark incorrectly because it extracts data from news reports rather than official filings, leading to data omission.
%     }
%     \label{fig:case3}
% \end{figure}

\begin{QuestionCase}[label={prompt:basic},title={A web-based product successfully addresses a question of Time-Sensitive Data Fetching in \benchmark by leveraging financial plugins.}]

\textbf{Question}\\
walmart current stock price\\

\textbf{Response}\\
As of the U.S. market close on August 27, 2025, the stock price of Walmart Inc. (Ticker: WMT) was \$96.08, marking a daily gain of 0.03\% with a trading turnover of \$1.241 billion. The price represented a slight increase from the previous day's close of \$96.05. During the session, the stock reached an intraday high of \$96.32 and a low of \$95.60.

In terms of recent performance, Walmart's stock has declined by 6.33\% over the past five trading days, yet it still maintains a year-to-date (YTD) gain of 7.13\%. Analysts note that despite facing short-term pressure from tariff costs, the company has sustained its market competitiveness through supply chain optimization and its omnichannel strategy.\\

\textbf{Search Sources}
\begin{enumerate}[label=\arabic*.]
    \item tradersunion.com\\
    Walmart Inc. (WMT) Stock Price Forecast for 30 days: Today, Tomorrow, Next Week\\
    \textit{August 27, 2025 - The current price of Walmart Inc. (WMT) is US\$96.02. The analytical forecast indicates that the price of WMT, expected to reach US\$96.12 by the end of the week, could rise/fall to US\$99.35 by the end of the month.}

    \item tradersunion.com\\
    Walmart Inc. (WMT) Stock Price Forecast for 30 days: Today, Tomorrow, Next Week\\
    \textit{August 27, 2025 - The current price of Walmart Inc. (WMT) is US\$96.02. The analytical forecast indicates that the price of WMT is expected to reach US\$96.12 by the end of the week, and could rise/fall to US\$99.35 by the end of the month.}

    \item in.benzinga.com\\
    Walmart's Options: A Look at What the Big Money is Thinking - Walmart (NYSE:WMT) - Benzinga\\
    \textit{August 26, 2025 - Whales with a lot of money to spend have taken a noticeably bearish stance on Walmart. Looking at options history for Walmart WMT we detected 15 trades. If we consider the specifics of each trade, it is accurate to state that 33\% of the investors opened trades with bullish expectations and 46\% with bearish.}

    \item \textbf{Tencent Financial Industry Database}\\
    Walmart WMT.N\\
    \textit{August 28, 2025 - Real-time Stock Quote for Walmart (Code: WMT.N) in US Stocks: Data Updated: 2025-08-27 16:04:58; Latest Price: \$96.08; Previous Close: \$96.05; Today's Open: \$95.90; Day's High: \$96.32; Day's Low: \$95.60; Day's Change: 0.03\%; Day's Amplitude: 0.74\%; Turnover Rate: 0.16\%; Volume (shares): 12,925,656; Turnover (USD): 1,241,094,879.00; Volume Ratio: 0.45; 5-Day Change: -6.33\%; YTD Change: 7.13\%; P/E Ratio (TTM): 36.26; P/B Ratio: 8.51; Dividend Yield: 0.98\%; 20-Day Change: -1.32\%; Total Market Cap (billion USD): 766.76.}

    \item tradersunion.com\\
    Walmart Inc. (WMT) Stock Price Forecast (30 days): Today, Tomorrow, Next Week\\
    \textit{August 27, 2025 - The current price of Walmart Inc. (WMT) is \$96.02. According to analytical forecasts, the price of WMT is expected to reach \$96.12 by the end of this week, and could rise/fall to \$99.35 by the end of this month.}

    \item marketbeat.com\\
    Walmart (NYSE:WMT) Price Target Raised to \$114.00\\
    \textit{August 24, 2025 - Walmart (NYSE:WMT) Price Target Raised to \$114.00 Key Points - The Goldman Sachs Group has raised Walmart's price target from \$101.00 to \$114.00, indicating a potential upside of 17.25\% from its previous close.}

    \item Tonghuashun Finance\\
    Walmart's turnover on August 27 was \$1.241 billion, a decrease of 41.91\% from the previous trading day.\\
    \textit{August 28, 2025 - On August 27, 2025, consumer staples retailer Walmart (WMT) had a turnover of \$1.241 billion, ranking 44th in US stocks for the day. The turnover decreased by 41.91\% from the previous day, and the trading volume was 12.9257 million shares.}

    \item WeChat Official Account · CC Finance\\
    Walmart, with Unimpressive Revenue and Profits, Has a Stock Price Stronger Than the Magnificent Seven\\
    \textit{July 29, 2025 - The continuous rise of Walmart's stock price (around \$95.67/share in July 2025, market cap of \$760 billion, up over 300\% from its 2020 low), despite single-digit growth in revenue and profit...}

    \item WeChat Official Account · Haishi Shibei\\
    Walmart (WMT) Stock Price Under Pressure! Profit Misses Expectations for the First Time in Three Years, Soaring Tariff Costs Drag Down Profit Outlook\\
    \textit{August 22, 2025 - Transcript from the conference call: "Good morning/afternoon, everyone. Welcome to Walmart's Fiscal Year 2026 Second Quarter Earnings Conference Call. All participants will now be in a listen-only mode..."}
\end{enumerate}
\end{QuestionCase}

\begin{QuestionCase}[label={prompt:basic},title={A web-based product fails a Single Historical Data Retrieval question in \benchmark due to the absence of search tool utilization.}]

\textbf{Question}\\
How much cash was used in investing activities by Apple in the 2021 fiscal year? (Answer in millions of dollars, rounded to nearest integer)\\

\textbf{Reasoning}\\
The user needs to know the amount of cash used in investing activities by Apple in the 2021 fiscal year, in millions of dollars, rounded to the nearest integer. First, I need to recall Apple's financial reports, specifically the cash flow statement for the 2021 fiscal year.

Apple's fiscal year typically ends in September, so the 2021 fiscal year would be the year ending September 2021. According to public financial data, in Apple's 2021 cash flow statement, the net cash flow from investing activities is usually negative because investment expenditures exceed income. The "cash used" here refers to the net amount of total cash outflows from investing activities minus inflows, but typically "cash used" in a question refers to the net cash outflow from investing activities.

Consulting Apple's 2021 10-K report, the investing activities section includes capital expenditures (like PP\&E), acquisitions, investment securities, etc. \textbf{From memory}, was the net cash outflow from investing activities for Apple's 2021 fiscal year approximately 144,266 million dollars? Or is more accurate data needed.

Confirming again, in Apple's 2021 fiscal year cash flow statement, the net cash flow from investing activities was -144,266 million dollars. This means the cash used for investing activities was 144,266 million dollars. So the answer should be 144266.\\

\textbf{Response}\\
144266\\

\textbf{Search Sources}\\
\textbf{No search sources were used.}

\end{QuestionCase}
\begin{QuestionCase}[label={prompt:basic},title={A web-based product successfully solves a Complex Historical Investigation question by scoping its search to a reliable source (nasdaq.com) and then cross-verifying the retrieved data through reverse searches.}]

\textbf{Question}\\
What was the change in opening price compared to the previous closing price (in dollars, unadjusted) on the effective date of Apple Inc.'s (AAPL) most recent stock split? What was the split ratio? (Round to two decimal places)\\

\textbf{Reasoning}\\
...\\
Now, let's try searching NASDAQ for "AAPL closing price August 28 2020" to confirm the exact figure.

\begin{itemize}
    \item \textbf{Searched for "AAPL closing price August 28 2020 site:nasdaq.com"}
        \begin{itemize}
            \item \textit{nasdaq.com} (x10)
        \end{itemize}

    \item \textbf{Searched for "AAPL opening price August 31 2020 site:nasdaq.com"}
        \begin{itemize}
            \item \textit{nasdaq.com} (x10)
        \end{itemize}
\end{itemize}
...\\
\textbf{Confirming closing price}\\
I'm searching for "AAPL August 28 2020 close 499.23" to verify the exact figure.

\begin{itemize}
    \item \textbf{Searched for "AAPL August 28 2020 close 499.23"}
        \begin{itemize}
            \item \textit{finance.yahoo.com}
            \item \textit{statmuse.com}
            \item \textit{stockanalysis.com}
            \item \textit{investor.apple.com}
            \item \textit{financecharts.com}
            \item \textit{fool.com}
        \end{itemize}

    \item \textbf{Searched for "AAPL August 31 2020 open 127.58"}
\end{itemize}
...\\

\textbf{Response}\\
The most recent stock split for Apple Inc. (AAPL) was effective on August 31, 2020, with a split ratio of 4-for-1.

On that date, the opening price was \$127.58, compared to the previous closing price of \$499.23 on August 28, 2020. The change was -\$371.65.\\

\end{QuestionCase}

\end{document}